\documentclass{article} %
\usepackage{iclr2027_conference,times}

\usepackage[utf8]{inputenc}
\usepackage[T1]{fontenc}
\usepackage{hyperref}
\usepackage{url}
\usepackage{booktabs}
\usepackage{amsfonts}
\usepackage{amsmath}
\usepackage{amssymb}
\usepackage{amsthm}
\usepackage{bbm}
\usepackage{graphicx}
\usepackage{xcolor}
\usepackage{microtype}
\usepackage{enumitem}
\usepackage{subcaption}
\usepackage{chngcntr}

\graphicspath{{figures/}{./}}

\title{ResBM: Residual Bottleneck Models for \\ Low-Bandwidth Pipeline Parallelism}

\author{%
Alan Aboudib$^{1,*,\dagger}$ \quad
Kalei Brady$^{1,*}$ \quad
Rodrigo Lopez Portillo A.$^{1}$ \quad
Felix Quinque$^{1}$ \\
{\bf Brian McCrindle$^{1}$ \quad
Steffen Cruz$^{1}$} \\[1ex]
$^{1}$Macrocosmos AI \\
\texttt{\{firstname\}@macrocosmos.ai} \\[1ex]
{\mdseries\small $^{*}$Core contributors \quad $^{\dagger}$Corresponding author}
}

\iclrfinalcopy

\newif\ifanon
\anontrue

\begin{document}

\maketitle
\lhead{Preprint}

\begin{abstract}
Unlocking large-scale, low-bandwidth decentralized training has the potential to utilize otherwise untapped compute resources.
In centralized settings, large-scale multi-node training is primarily enabled by data and pipeline parallelism, two techniques that typically rely on high-bandwidth communication.
While efficient methods now exist for decentralized data parallelism, pipeline parallelism remains a major challenge.
Recent efforts, such as Subspace Models (SM), report up to $100\times$ activation compression but must constrain the embedding and every transformer block before the last pipeline boundary to a shared low-rank subspace.
Bottleneck layers, as in SWARM parallelism, compress only at pipeline boundaries but interrupt the residual identity path, degrading training at aggressive compression rates or with many boundaries.
We propose the Residual Bottleneck Model (ResBM), which combines the strengths of both: a residual encoder--decoder bottleneck applied only at pipeline boundaries, trained end-to-end with standard optimizers, that preserves an explicit low-rank identity path across them.
In our main comparative experiments with models of up to 8B parameters and training runs of up to $11$B tokens on large-scale web datasets, we show that ResBM achieves the lowest loss among evaluated methods at extreme compression rates of $100\times$ and $128\times$.
Although the authors of SM have reported little or no degradation from activation compression, our longer-horizon, optimizer-matched evaluations reveal a measurable loss gap for every compression method evaluated, including ResBM.
However, ResBM consistently exhibits the smallest such gap relative to uncompressed baselines.
Its loss gap remains approximately constant during later training, whereas SM's appears to grow over the evaluated horizon.
We therefore position ResBM as a new state of the art for extreme activation compression in low-bandwidth pipeline-parallel training.

\end{abstract}

\section{Introduction}
\label{sec:introduction}

Training frontier large language models (LLMs) requires distributing computation across hundreds or thousands of accelerators.
As models grow beyond the memory capacity of individual devices, two complementary parallelism strategies are commonly combined: data parallelism (DP), which replicates the model across workers that process different data shards, and pipeline parallelism (PP), which partitions the model's layers across devices and streams activations forward and gradients backward \citep{narayanan2021megatron, huang2019gpipe}.
Both strategies depend on high-bandwidth interconnects, NVLink for intra-node communication and InfiniBand for inter-node transfers, 
making conventional large-scale training heavily dependent on specialized, expensive datacenter infrastructure.

The high cost and limited availability of such clusters have motivated interest in decentralized training, where geographically distributed devices connected over the public internet collaborate to train a single model \citep{NEURIPS2021_41a60377, yuan2022tasklets, ryabinin2023swarmparallelismtraininglarge}.
While low-communication data parallelism is now relatively mature, with methods such as DeMo \citep{peng2026iclr-demo}, DiLoCo and its variants \citep{douillard2023diloco, douillard2025streamingdilocooverlappingcommunication, sarfi2026communication} reducing DP synchronization overhead by up to $500\times$, pipeline parallelism remains the primary bottleneck.
In datacenters, PP communication traverses InfiniBand links at $400$\,Gbps per port \citep{nvidia2026quantum2}, whereas typical consumer upload bandwidth is around $60$\,Mbps \citep{epoch2025howfarcandecentralizedtrainingovertheinternetscale}, so naively matching datacenter transfer times would require compressing activations and their gradients by three to four orders of magnitude.
However, as \citet{ryabinin2023swarmparallelismtraininglarge} note, the effective requirement is far lower than this naive ratio for larger models: by the \emph{square-cube law}, compute grows faster than communication with model width, leaving more room to overlap the two.

Compressing activations for pipeline parallelism is harder than compressing weight gradients for data parallelism: activations propagate sequentially through layers, so compression errors accumulate with depth. Simple strategies such as 8-bit quantization save little, while the bottleneck layers of SWARM parallelism \citep{ryabinin2023swarmparallelismtraininglarge}, which compress the full block output at each pipeline boundary, degrade convergence at aggressive compression rates (Section~\ref{sec:compress_local_perf}).
This suggests that where compression is applied matters as much as its magnitude: in particular, placing learned or lossy transformations on the residual identity path, the connection that makes deep networks trainable \citep{he2016deep}, breaks the identity mapping and risks destabilizing optimization.

Recently, Subspace Models (SM) were proposed by \citet{ramasinghe2025neurips-subspace}, claiming to achieve up to $100\times$ activation compression with little or no degradation. Their approach constrains the embedding and the out-projection matrices of every transformer block before the last pipeline boundary to a shared low-rank subspace: out-projection matrices remain trainable but are restricted to this subspace, while the embedding is decomposed into trainable low-rank and fixed high-rank components, allowing low-rank activations to be transmitted and reconstructed at the receiving node. This constraint applies whether or not a block precedes a boundary, so only the blocks of the last stage operate at full rank, restricting the effective parameter space available to the model. Moreover, these results were established over a relatively short training horizon. Our longer experiments suggest that the cost of these constraints becomes visible only later in training (Section~\ref{sec:compress_local_perf}).

In this paper, we propose the Residual Bottleneck Model (ResBM), an architecture designed specifically for low-bandwidth pipeline parallelism that combines the strengths of SM and of SWARM's bottleneck layers. Like SWARM and unlike SM, ResBM compresses only at pipeline boundaries: it adds one encoder--decoder pair per boundary, trained end-to-end with the rest of the model, and leaves every transformer block unconstrained. However, unlike SWARM, it preserves a partial identity path across every boundary: fixed, parameter-free projections carry an exact low-rank identity that bypasses each encoder and decoder, so that no boundary breaks the identity mapping. We relax the lossless-compression assumption and instead measure the degradation that extreme activation compression induces over extended training, across model sizes, optimizers, compression rates, and numbers of pipeline stages. Across these settings, ResBM consistently exhibits the smallest degradation among the methods we evaluate. In our 2B and 3.5B experiments with eight or more pipeline stages, its loss gap to the uncompressed baseline stays at about 0.24--0.30 nats after the initial phase of training, whereas SM's grows to 0.40--0.45 nats. ResBM also outperforms SM on a reproduction of a public 8B-parameter SM run whose pipeline stages are sized to consumer and workstation GPUs \citep{avraham2026agora}, and, trained with Muon, reaches the loss of the uncompressed AdamW baseline within 26B tokens. These results establish ResBM as the new state of the art for extreme activation compression in low-bandwidth pipeline-parallel training.

\section{Related Work}
\label{sec:related_work}

\paragraph{Decentralized training and activation compression.}

Decentralized training distributes computation across autonomous, geographically dispersed devices connected by heterogeneous networks \citep{NEURIPS2021_41a60377}. Systems such as SWARM parallelism \citep{ryabinin2023swarmparallelismtraininglarge} and Tasklets \citep{yuan2022tasklets} address scheduling and resource allocation in such environments. SWARM additionally explores compression-aware architectures using 8-bit quantization or bottleneck layers applied to the full block output, including the residual stream, but only at modest $2$--$4\times$ compression. We use this bottleneck as an ablation baseline, where adding ResBM's residual identity path isolates its architectural contribution (Section~\ref{sec:compress_local_perf}).

Activation compression for pipeline parallelism is particularly challenging because lossy errors propagate across layers and can accumulate with model depth \citep{bian2024compressing}, while activation distributions also vary substantially across samples and batches \citep{rudakov2023activations}. Standard approaches such as top-$k$ sparsification, SVD projection, and quantization therefore degrade or fail at aggressive compression rates \citep{ramasinghe2025neurips-subspace}. Subspace Models (SM) address this by constraining embedding and out-projection matrices to a shared low-rank subspace, claiming up to $100\times$ compression with little or no degradation. 
Every block before the last pipeline boundary must operate in this subspace, whether or not it precedes a boundary, restricting the model's effective parameter space; ResBM instead adds one encoder--decoder pair per boundary and leaves every transformer block unconstrained.
Using the authors' released SM implementation from Agora \citep{avraham2026agora}, we find that over longer training horizons SM develops a loss gap relative to the uncompressed baseline that grows with tokens seen. ResBM's gap instead remains approximately stable after the initial phase of training (Section~\ref{sec:compress_local_perf}).

\paragraph{Residual connections and identity preservation.}
The residual connection, introduced by \citet{he2016deep}, is a cornerstone of modern deep network design.
By adding an identity shortcut around each stack of layers, it addresses the degradation problem of very deep plain networks, making them easier to optimize and enabling training of substantially deeper architectures.
In transformer-based LLMs, the residual stream carries information across all layers and is essential for stable optimization.
Recent work by \citet{xie2026mhcmanifoldconstrainedhyperconnections} has shown that some transformations can be safely introduced along the residual path provided they preserve the identity mapping property of the connection.
Our architecture builds on this insight: the bottleneck projections are part of the residual functions, the encoder following the feed-forward layer and the decoder preceding attention, while a low-rank identity path bypasses each and remains uninterrupted even as the communicated activation dimensionality is reduced.

\section{Residual Bottleneck Models}
\label{sec:resbm}

\subsection{Notation}
\label{sec:notation}

Modern LLMs are typically implemented as a stack of transformer blocks \citep{NIPS2017_3f5ee243}, each built around self-attention and the residual connection.
A single residual sublayer of such a block can be written as
\begin{equation}
    x_{i+1} = x_i + F(x_i, W_i),
\end{equation}
where $x_i \in \mathbb{R}^{L \times H}$ denotes the input to residual sublayer $i$, with $L$ the sequence length and $H$ the hidden dimension.
The residual function $F(\cdot, W_i)$ typically corresponds to either a self-attention or a feed-forward sublayer parameterized by weights $W_i$; each such sublayer performs one residual addition and therefore advances the index by one, so a transformer block occupies two consecutive sublayer indices.
Activations are represented with one token per row, so a linear layer with weight matrix $W$ acts by right multiplication, $y = xW$, and its outputs lie in the row space of $W$.

More generally, if the hidden dimensionality is allowed to change from one sublayer to the next, the residual connection must account for this mismatch.
Let $N$ denote the number of residual sublayers and let $c_i$ denote the width of the residual stream entering sublayer $i$, so that $x_i \in \mathbb{R}^{L \times c_i}$ and $F(x_i, W_i) \in \mathbb{R}^{L \times c_{i+1}}$.

The identity branch must be mapped to the width of the sublayer output before the two can be added.
We do this with a fixed, parameter-free projection built from rectangular identity matrices.
For integers $p, q \ge 1$ and $0 \le k \le \min(p, q)$, let $\mathbf{I}^{(k)}_{p \times q} \in \mathbb{R}^{p \times q}$ denote the rank-$k$ rectangular identity matrix defined elementwise as
\begin{equation}
\label{eq:rect_identity_definition}
\left[\mathbf{I}^{(k)}_{p \times q}\right]_{a,b} =
\begin{cases}
1, & \text{if } a = b \text{ and } a \le k, \\
0, & \text{otherwise},
\end{cases}
\end{equation}
and write $\mathbf{I}_{p \times q} \mathrel{:=} \mathbf{I}^{(\min(p,q))}_{p \times q}$ for the full-rank case, so that $\mathbf{I}_{p \times p}$ is the usual identity matrix $\mathbf{I}_p$.
The \emph{identity projection operator} of sublayer $i$ is the matrix
\begin{equation}
\label{eq:Pid_definition}
\mathcal{P}^{\mathrm{id}}_i
\mathrel{:=}
\mathbf{I}_{c_i \times c_{i+1}}
\in \mathbb{R}^{c_i \times c_{i+1}},
\qquad\text{so that}\qquad
x_i \mathcal{P}^{\mathrm{id}}_i \in \mathbb{R}^{L \times c_{i+1}}
\quad\text{for}\quad
x_i \in \mathbb{R}^{L \times c_i},
\end{equation}
acting on the identity branch by right-multiplication.
If $c_i < c_{i+1}$ it zero-pads the identity branch, appending $c_{i+1}-c_i$ zero columns; if $c_i > c_{i+1}$ it truncates the last $c_i-c_{i+1}$ columns; and if $c_i = c_{i+1}$ it reduces to the standard identity matrix $\mathbf{I}_{c_i}$.
The standard transformer is recovered by setting $c_i = H$ for all $i$.
The generalized residual update becomes
\begin{equation}
\label{eq:generalized_transformer_equation}
    x_{i+1} = x_i \mathcal{P}^{\mathrm{id}}_i + F(x_i, W_i),
\end{equation}
and unrolling it from any sublayer $m$ to any later sublayer $n$, with $0 \le m < n \le N$, gives
\begin{equation}
\label{eq:recursive_transformer_equation}
    x_n =  x_m \left( \prod_{i=m}^{n-1} \mathcal{P}^{\mathrm{id}}_{i} \right) + \sum_{i=m}^{n-1} F(x_i, W_i) \left( \prod_{j=i+1}^{n-1} \mathcal{P}^{\mathrm{id}}_{j} \right),
\end{equation}
where products of matrices are taken in order of increasing index from left to right, and an empty product denotes the identity matrix.
Section~\ref{sec:identity_path} shows that these projections preserve the identity mapping property of the residual connection.

\subsection{The Bottleneck Layer}
\label{sec:bottleneck_layer}

A bottleneck layer is introduced at every pipeline boundary, after the feed-forward layer of the last block before it; the output of the model's final block is consumed locally and never crosses a boundary.
It consists of an encoder, which maps the feed-forward output to a lower-dimensional bottleneck space before the boundary, and a decoder, which maps the received state back to the working width after it; each is a two-layer bias-free MLP with an elementwise nonlinearity after each linear layer.

\paragraph{Setting.}
Consider a pipeline boundary and let $i$ be the index, in the sublayer indexing of Section~\ref{sec:notation}, of the feed-forward sublayer of the block before the boundary, so that $i+1$ is the attention sublayer of the block after it.
Writing $h$ for the bottleneck width, the widths are $c_i = H$, $c_{i+1} = h$ and $c_{i+2} = H$, so the residual stream narrows to $h$ at each boundary and returns to $H$ after it, and by Equation~\ref{eq:Pid_definition} the identity projections at the two sublayers are
\[
\mathcal{P}^{\mathrm{id}}_i = \mathbf{I}_{H \times h},
\qquad
\mathcal{P}^{\mathrm{id}}_{i+1} = \mathbf{I}_{h \times H},
\]
the first truncating the identity branch to its leading $h$ coordinates and the second zero-padding it back to width $H$.
All other sublayers, which are not adjacent to a boundary, have equal input and output widths and the standard identity $\mathbf{I}_H$.
$\mathrm{Attn}_i$ and $\mathrm{FF}_i$ denote the complete attention and feed-forward sublayer maps, including their post-normalization, and $W^{\mathrm{Attn}}_i$ and $W^{\mathrm{FF}}_i$ their parameters.

\paragraph{Encoder and transmitted state.}
Let $x_i \in \mathbb{R}^{L \times H}$ be the post-attention state entering the feed-forward sublayer $i$, and $z_i = \mathrm{FF}_i(x_i) \in \mathbb{R}^{L \times H}$ the feed-forward output, immediately before compression.
The encoder $E_i$, with parameters $W^{E}_i$, maps it to the \emph{compressed residual contribution}
\[
b_i = E_i(z_i), \qquad b_i \in \mathbb{R}^{L \times h}, \qquad h < H .
\]
Concretely, writing $\phi$ for the elementwise non-linearity and $m$ for the intermediate width of the encoder and decoder,
\begin{equation}
\label{eq:encoder_decoder_shapes}
    E_i(z_i) = \phi\bigl(\phi\bigl(z_i\,W^{E}_{i,1}\bigr)\,W^{E}_{i,2}\bigr),
    \qquad W^{E}_{i,1} \in \mathbb{R}^{H \times m},
    \quad W^{E}_{i,2} \in \mathbb{R}^{m \times h},
    \qquad m = H/4,
\end{equation}
so that $W^{E}_i = \{W^{E}_{i,1}, W^{E}_{i,2}\}$ and the encoder maps $H \to m \to h$.

Because $E_i$ is part of the residual function of sublayer $i$, its output is summed with the projected identity branch before transmission, and the tensor sent across the boundary is the \emph{compressed residual state}
\begin{equation}
\label{eq:compressed_residual_state}
    x_{i+1} = x_i \mathcal{P}^{\mathrm{id}}_i + b_i,
    \qquad x_{i+1} \in \mathbb{R}^{L \times h},
\end{equation}
which is Equation~\ref{eq:generalized_transformer_equation} at sublayer $i$ with residual function $F(\cdot, \{W^{\mathrm{FF}}_i, W^{E}_i\}) = E_i \circ \mathrm{FF}_i$; the encoder parameters belong to the residual function but not to $\mathrm{FF}_i$, so the parameter set of the sublayer is the union of the two.

\paragraph{Decoder and expanded state.}
On the receiving stage, the decoder $D_{i+1}$, with parameters $W^{D}_{i+1}$, acts on the \emph{received state} $x_{i+1}$:
\begin{equation}
\label{eq:decoder_definition}
    \tilde{x}_{i+1} = D_{i+1}(x_{i+1}), \qquad \tilde{x}_{i+1} \in \mathbb{R}^{L \times H},
\end{equation}
with the decoder being the mirror image of the encoder:
\begin{equation}
\label{eq:decoder_mlp}
    D_{i+1}(x_{i+1}) = \phi\bigl(\phi\bigl(x_{i+1}\,W^{D}_{i+1,1}\bigr)\,W^{D}_{i+1,2}\bigr),
    \qquad W^{D}_{i+1,1} \in \mathbb{R}^{h \times m},
    \quad W^{D}_{i+1,2} \in \mathbb{R}^{m \times H},
\end{equation}
so that $W^{D}_{i+1} = \{W^{D}_{i+1,1}, W^{D}_{i+1,2}\}$ and the decoder maps $h \to m \to H$. The decoder output $\tilde{x}_{i+1}$ is the tensor that enters the attention map of sublayer $i+1$.
Because $x_{i+1}$ already carries the projected identity branch, $\tilde{x}_{i+1}$ is an expansion of the compressed residual state rather than a reconstruction of $z_i$: despite its encoder--decoder structure, the bottleneck is not an autoencoder and is not trained under a reconstruction objective.
Equation~\ref{eq:generalized_transformer_equation} at sublayer $i+1$, with residual function $F(\cdot, \{W^{\mathrm{Attn}}_{i+1}, W^{D}_{i+1}\}) = \mathrm{Attn}_{i+1} \circ D_{i+1}$, restores the working width:
\begin{equation}
\label{eq:expanded_residual_state}
    x_{i+2} = x_{i+1} \mathcal{P}^{\mathrm{id}}_{i+1} + \mathrm{Attn}_{i+1}(\tilde{x}_{i+1}),
    \qquad x_{i+2} \in \mathbb{R}^{L \times H}.
\end{equation}
The state $x_{i+2}$ enters the feed-forward sublayer $i+2$ of the receiving block; if that block also precedes a boundary, Equation~\ref{eq:compressed_residual_state} applies to it in turn.

Appendix Figure~\ref{fig:app:bottleneck_layer} shows the resulting layout: the encoder ends the sending block, the decoder starts the receiving one, communication scales with $h$ rather than $H$, and the identity branch bypasses both, acted on only by the fixed projections $\mathcal{P}^{\mathrm{id}}_i$ and $\mathcal{P}^{\mathrm{id}}_{i+1}$, which appear in the figure as splits and concatenations of the residual stream.

\subsection{The identity path}
\label{sec:identity_path}

Building on the rank-collapse motivation of \citet{ramasinghe2025neurips-subspace} and the output-projection spectra in Appendix~\ref{app:baseline_spectra}, we investigate whether the identity path can likewise be narrowed. If its forward activations and backward gradients are sufficiently concentrated, this may permit lower-dimensional communication with limited impact on optimization.
To retain a direct shortcut through repeated width changes, we choose the fixed projections $\mathcal{P}^{\mathrm{id}}_i$ of Equation~\ref{eq:Pid_definition} to preserve an exact identity mapping on the communicated subspace.

Substituting Equation~\ref{eq:Pid_definition} into Equation~\ref{eq:recursive_transformer_equation}, every product of identity projections telescopes into a single projection: for any sequence of widths $c_0, c_1, \dots, c_N$ and any $0 \le m < n \le N$,
\begin{equation}
\label{eq:identity_composition}
    \prod_{i=m}^{n-1} \mathbf{I}_{c_i \times c_{i+1}} = \mathbf{I}^{(k)}_{c_m \times c_n},
    \qquad
    k = \min_{m \le i \le n} c_i .
\end{equation}
The composition is the $c_m \times c_n$ rectangular identity truncated to rank $k$, the narrowest width anywhere along the path: a coordinate survives the whole composition precisely when it is present at every intermediate width.
For the widths of Section~\ref{sec:bottleneck_layer}, sublayers within a stage contribute $\mathbf{I}_H$ and leave the product unchanged. Measured from one communication boundary to the next, the composition is therefore
\[
\mathbf{I}_{h \times H}\,\mathbf{I}_{H \times h} = \mathbf{I}_{h},
\]
a full-rank identity on the $h$-dimensional communicated stream: that stream carries an exact, unbroken identity path across any number of stages.
Measured in the $H$-dimensional working width, each boundary contributes
\[
\mathbf{I}_{H \times h}\,\mathbf{I}_{h \times H} = \mathbf{I}^{(h)}_{H \times H},
\]
so any path crossing at least one communication boundary has $k = h$, independently of its length.

This construction preserves an exact identity shortcut on the communicated $h$-dimensional subspace, providing a direct gradient contribution that bypasses the learned encoder and decoder transformations.
These coordinates have no predefined semantic interpretation; end-to-end training can adapt the surrounding representations to route the most expressed directions through the preserved shortcut.

\section{Experiments}
\label{sec:experiments}

\subsection{Experimental setup}
\label{sec:experiment_setup}

Unless otherwise stated, all models are trained on the C4 dataset \citep{raffel2023exploringlimitstransferlearning} with a batch size of 32, using AdamW~\citep{DBLP:conf/iclr/LoshchilovH19} with a peak learning rate of $3\times10^{-4}$ and a weight decay of $0.01$. The learning rate warms up linearly from $1\%$ of its peak over 500 steps, then follows a cosine decay to $1\%$ of its peak over 1T tokens, so that it stays within $0.2\%$ of its peak over our training horizons. All runs use fixed seeds, so that within each model scale every method trains on the same token stream in the same order, and both ResBM and SWARM start from the same initial weights as the uncompressed baseline; due to compute constraints, each configuration is trained with a single seed.

The uncompressed baseline, which also serves as the default backbone of every compressed model unless otherwise specified, is a transformer (approximately 2B parameters) based on the OLMo-2 architecture \citep{olmo20252olmo2furious} with reordered (RMS) normalization \citep{liu2022swine} and QK-Norm \citep{pmlr-v202-dehghani23a}. It has 8 transformer blocks, hidden width $H = 4096$, a 50K-token vocabulary, and a 1024-token context.

For ResBM, the encoder and decoder are two-layer bias-free MLPs (Equations~\ref{eq:encoder_decoder_shapes} and~\ref{eq:decoder_mlp}) with intermediate width $m = H/4$ and SiLU as the nonlinearity $\phi$. These choices were fixed a priori and not tuned; sweeping them, as well as potentially varying $h$ across boundaries, is left to future work, as it requires substantial compute resources.

For SWARM, we use the encoder and decoder of Equations~\ref{eq:encoder_decoder_shapes} and~\ref{eq:decoder_mlp}, consistent with the bottleneck design of \citet{ryabinin2023swarmparallelismtraininglarge}, so that its parameter count matches ResBM exactly.
As in the original design, the bottleneck compresses the full block output, residual stream included, $\tilde{x}_{i+1} = D_{i+1}\bigl(E_i(x_i + \mathrm{FF}_i(x_i))\bigr)$ and $x_{i+2} = \tilde{x}_{i+1} + \mathrm{Attn}_{i+1}(\tilde{x}_{i+1})$: the shortcut stops before the encoder and restarts after the decoder.
In ResBM, the identity branch instead bypasses the encoder and decoder (Equations~\ref{eq:compressed_residual_state} and~\ref{eq:expanded_residual_state}), so the rank-$h$ shortcut spans all boundaries uninterrupted (Equation~\ref{eq:identity_composition}).
SWARM thus serves as an ablation of ResBM's partial identity path, their only architectural difference.

For SM, we use the authors' Agora implementation \citep{avraham2026agora}, which enforces the output-projection subspace constraints through reparameterization, allowing the trainable parameters to be optimized with off-the-shelf optimizers such as AdamW and Muon~\citep{jordan2024muon}.
We evaluate two variants. The first, \textbf{SM}, uses exactly the same OLMo-2 backbone as ResBM, SWARM and the uncompressed baseline, with 8 transformer blocks.
Because SM restricts the output projections to a shared subspace, it trains fewer parameters than ResBM and SWARM.
For every comparison, we therefore also train a parameter-matched variant, \textbf{SM-M}, which adds transformer blocks until its parameter count matches ResBM's, distributing them to balance training FLOPs across pipeline stages (Appendix~\ref{app:sm_m}).

Unless otherwise specified, models are split into 8 pipeline stages, giving seven communication boundaries at which activations and their gradients are compressed by a factor equal to the specified compression rate. For ResBM, SWARM and SM, this places a boundary after each transformer block except the last; SM-M has more blocks, so some of its stages contain more than one block (Table~\ref{tab:sm_m_arch}).

All experiments run as a single pipeline replica on NVIDIA H100 GPUs, with one pipeline stage per GPU, so that the number of GPUs equals the number of stages. Pipeline parallelism is GPipe-style and implemented with PyTorch distributed over NCCL. For the compressed models, the inter-stage links are emulated: their bandwidth is capped at 80~Mbps, matching the low-bandwidth setting of \citet{ramasinghe2025neurips-subspace}, and a latency of 26~ms is added, so that all compressed runs share identical, reproducible network conditions. The uncompressed baseline instead communicates over a high-bandwidth 10~Gbps PCIe link. Unless otherwise noted, figures plot training loss against wall-clock time, with bars giving each run's average throughput in tokens per second. We define the loss gap as the difference, in nats, between the training loss of a compressed model and that of the uncompressed baseline trained with the same optimizer, at the same wall-clock time; it therefore also reflects any difference in throughput.
ResBM adds about 3.1\% to the parameters and 3.4\% to the per-token training compute of the 2B model, while dividing the traffic at each boundary by the compression rate; Appendix~\ref{app:appendix_throughput} details these overheads and reports throughput gains across bandwidths.

\subsection{Performance under extreme compression}
\label{sec:compress_local_perf}

\paragraph{Extreme compression rates.}
We first compare ResBM against SM \citep{ramasinghe2025neurips-subspace}, its parameter-matched variant SM-M, and SWARM \citep{ryabinin2023swarmparallelismtraininglarge} under extreme compression rates of $102.4\times$ (denoted $100\times$) and $128\times$, corresponding to bottleneck widths $h = 40$ and $h = 32$, respectively.

\begin{figure*}[t]
    \centering
    \begin{minipage}[c]{0.58\linewidth}
        \centering

        \begin{subfigure}[t]{\linewidth}
            \centering
            \includegraphics[width=\linewidth]{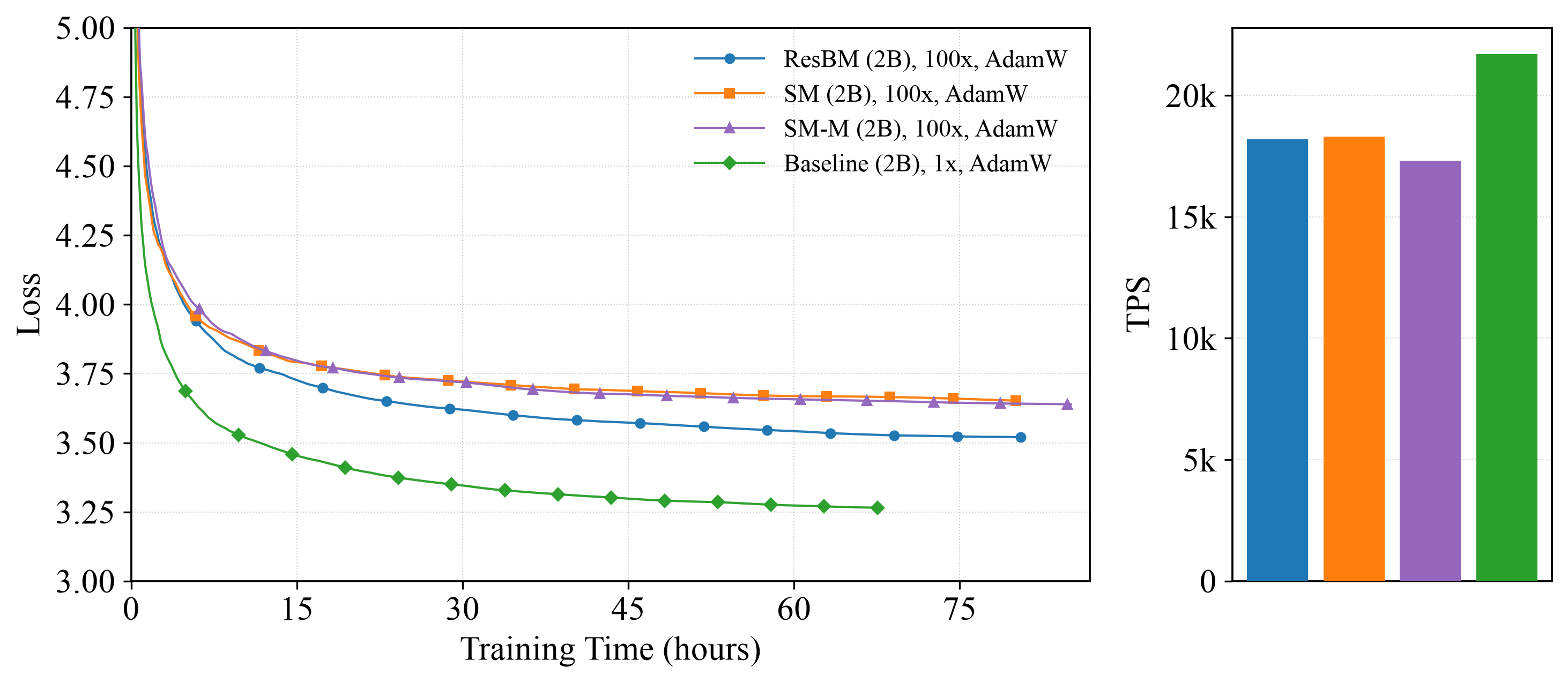}
            \caption{\small Training loss and throughput ($100\times$).}
            \label{fig:diff_100x_adamw_top}
        \end{subfigure}

        \par\smallskip

        \begin{subfigure}[t]{\linewidth}
            \centering
            \includegraphics[width=\linewidth]{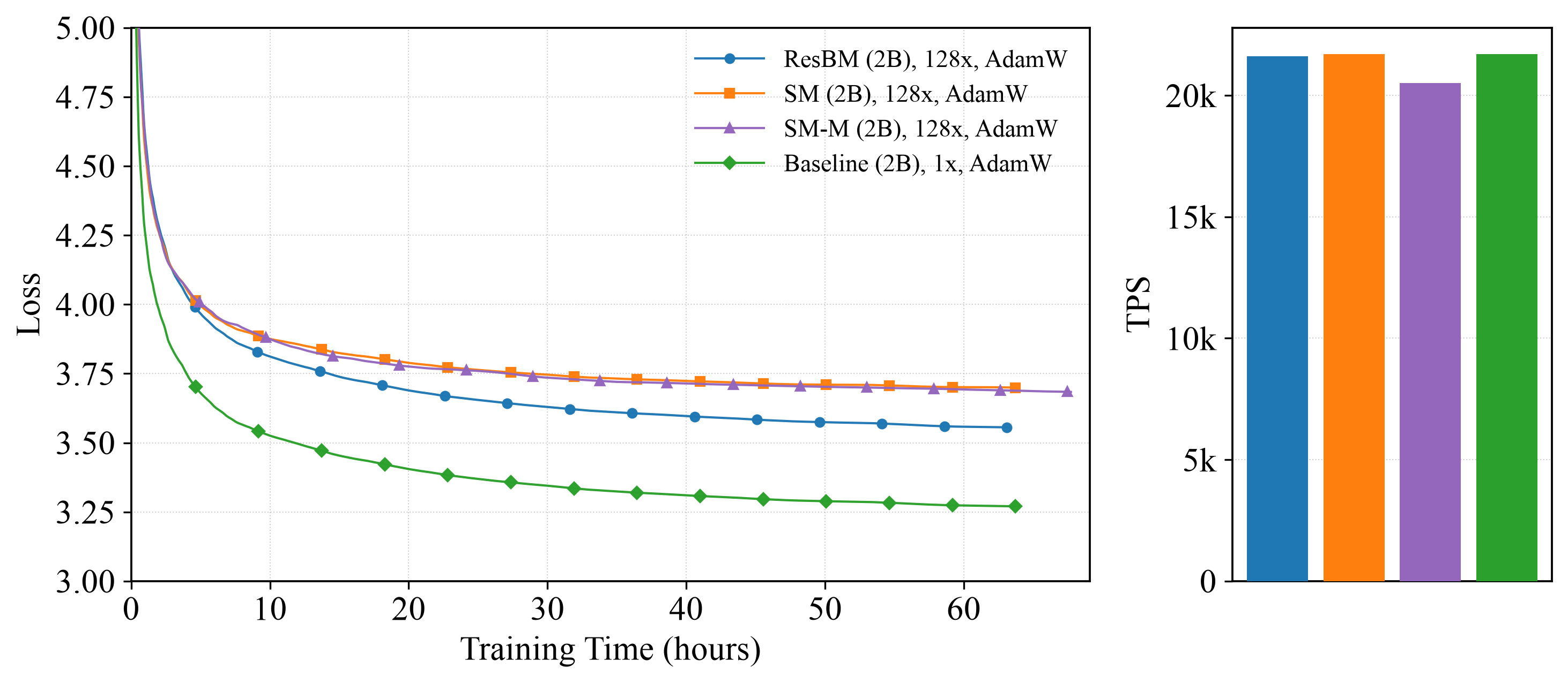}
            \caption{\small Training loss and throughput ($128\times$).}
       
            \label{fig:128x_adamw_mid}
        \end{subfigure}

    \end{minipage}
    \hfill
    \begin{minipage}[c]{0.41\linewidth}
        \centering

        \begin{subfigure}[t]{\linewidth}
            \centering
            \includegraphics[width=\linewidth]{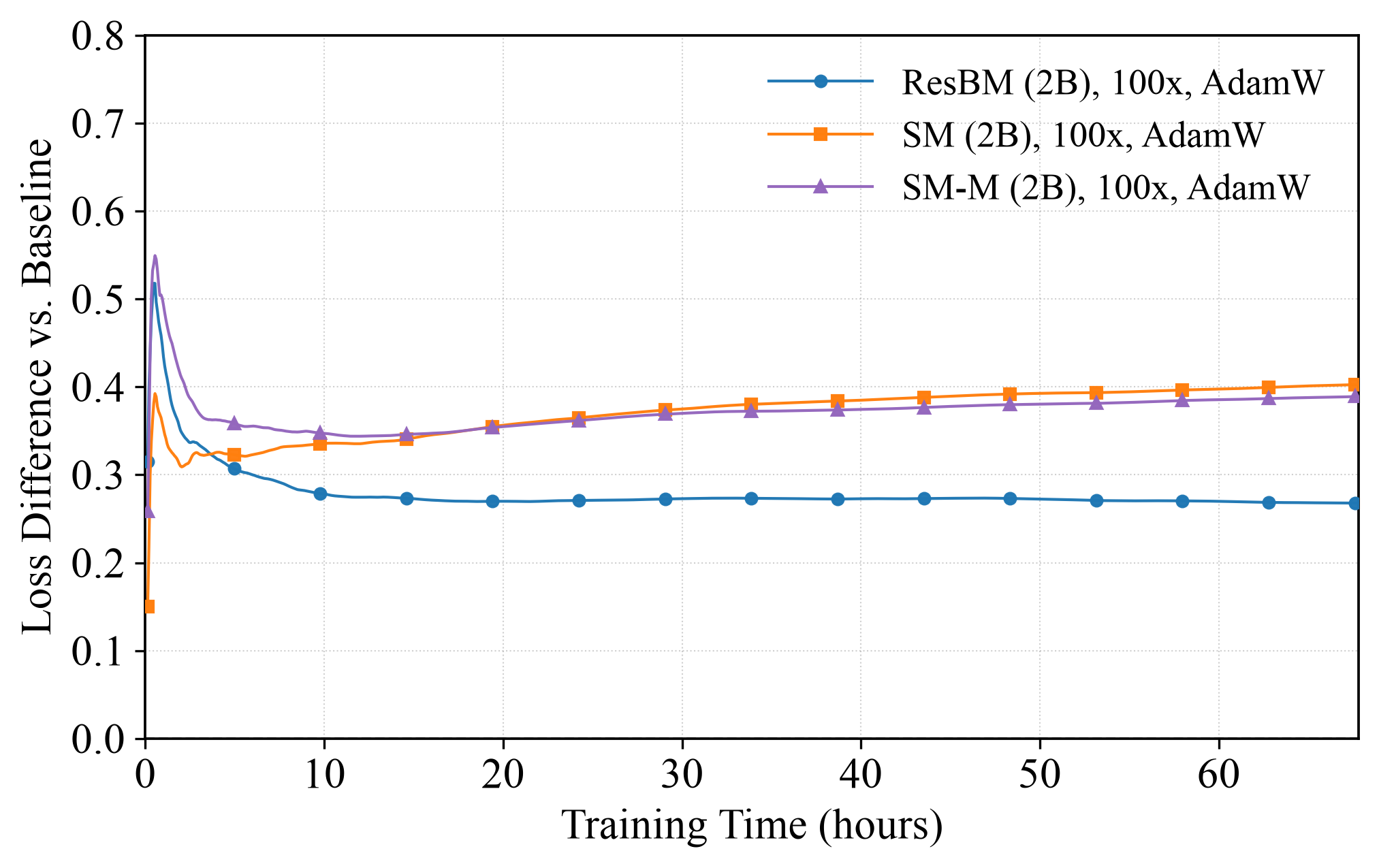}
            \caption{\small Evolution of loss gap (100$\times$).}
            \label{fig:diff_128x_adamw_top}
        \end{subfigure}

        \par\smallskip

        \begin{subfigure}[t]{\linewidth}
            \centering
            \includegraphics[width=\linewidth]{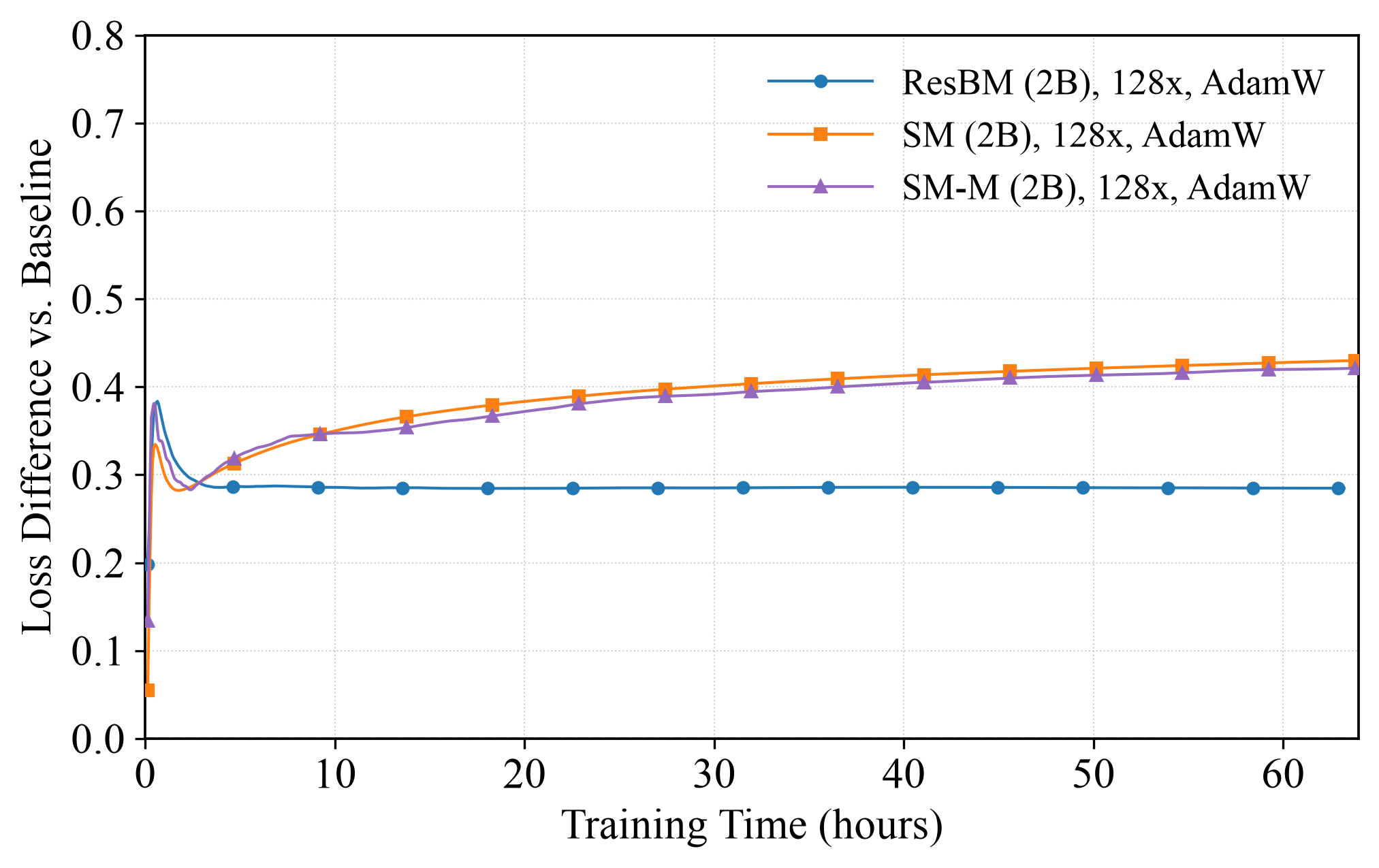}
            \caption{\small Evolution of loss gap (128$\times$).}
            \label{fig:diff_128x_adamw_bot}
        \end{subfigure}

    \end{minipage}
    \caption{
        \textbf{Training loss under extreme compression (AdamW).} ResBM, SM, and SM-M (2B parameters, about 5B tokens) at $100\times$ (a,c) and $128\times$ (b,d) compression. (c,d) Loss gap to the uncompressed AdamW baseline: ResBM's stays flat, while SM's grows.
    }

    \label{fig:resbm_vs_all_high_comp}
\end{figure*}

\begin{figure*}[t]
    \centering
    \begin{minipage}[c]{0.58\linewidth}
        \centering

        \begin{subfigure}[t]{\linewidth}
            \centering
            \includegraphics[width=\linewidth]{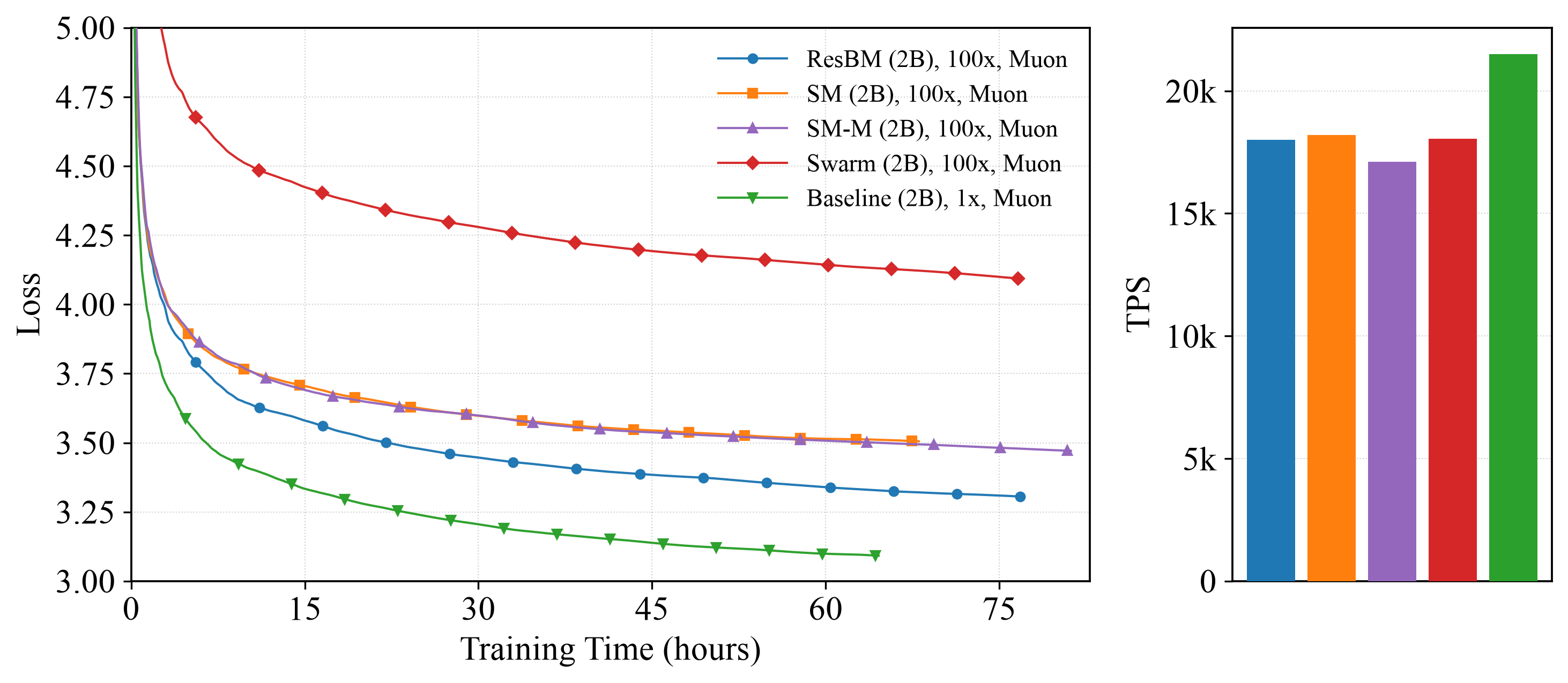}
            \caption{\small Training loss and throughput ($100\times$).}
            \label{fig:100x_muon}
        \end{subfigure}

        \par\smallskip

        \begin{subfigure}[t]{\linewidth}
            \centering
            \includegraphics[width=\linewidth]{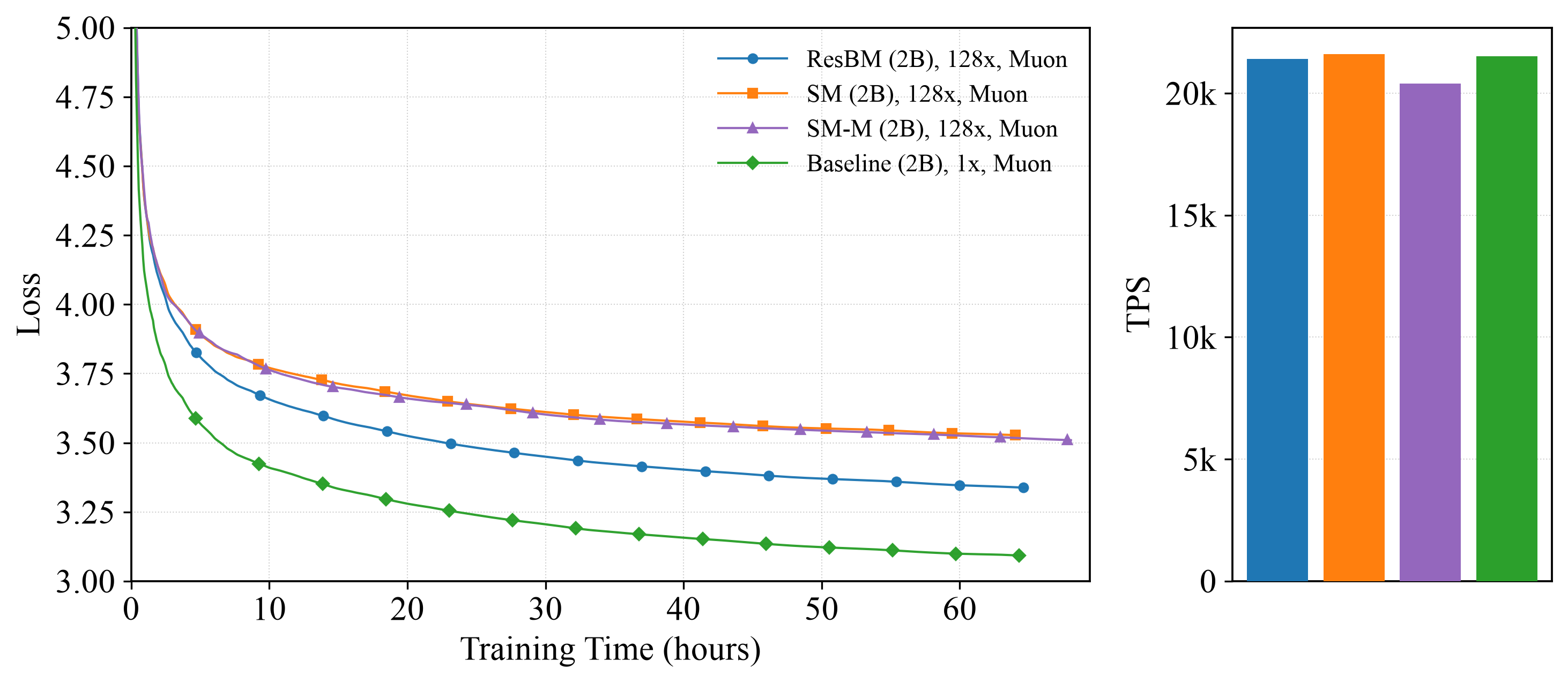}
            \caption{\small Training loss and throughput ($128\times$).}
            \label{fig:128x_muon}
        \end{subfigure}

    \end{minipage}
    \hfill
    \begin{minipage}[c]{0.41\linewidth}
        \centering

        \begin{subfigure}[t]{\linewidth}
            \centering
            \includegraphics[width=\linewidth]{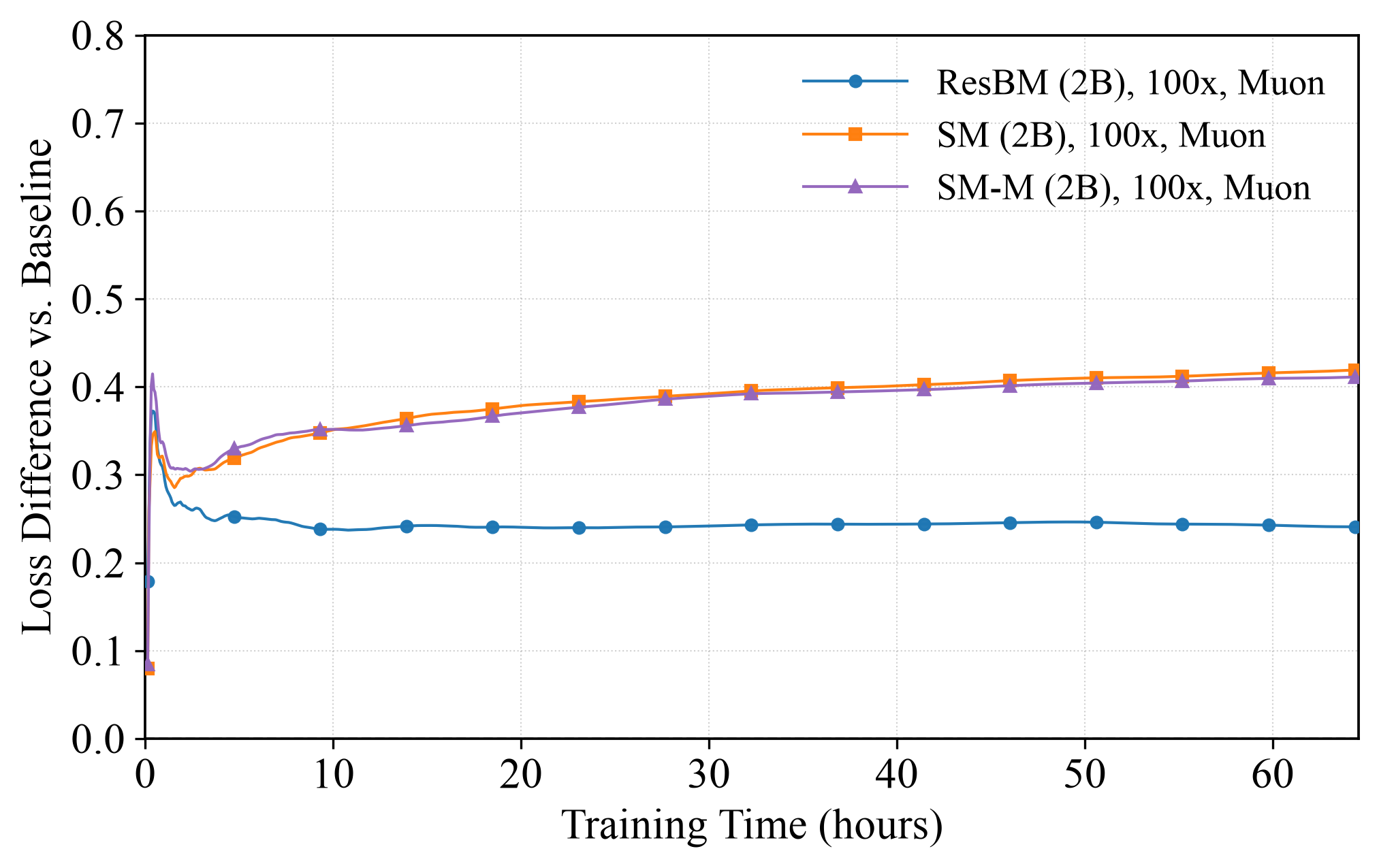}
            \caption{\small Evolution of loss gap (100$\times$).}
            \label{fig:diff_100x_muon}
        \end{subfigure}

        \par\smallskip

        \begin{subfigure}[t]{\linewidth}
            \centering
            \includegraphics[width=\linewidth]{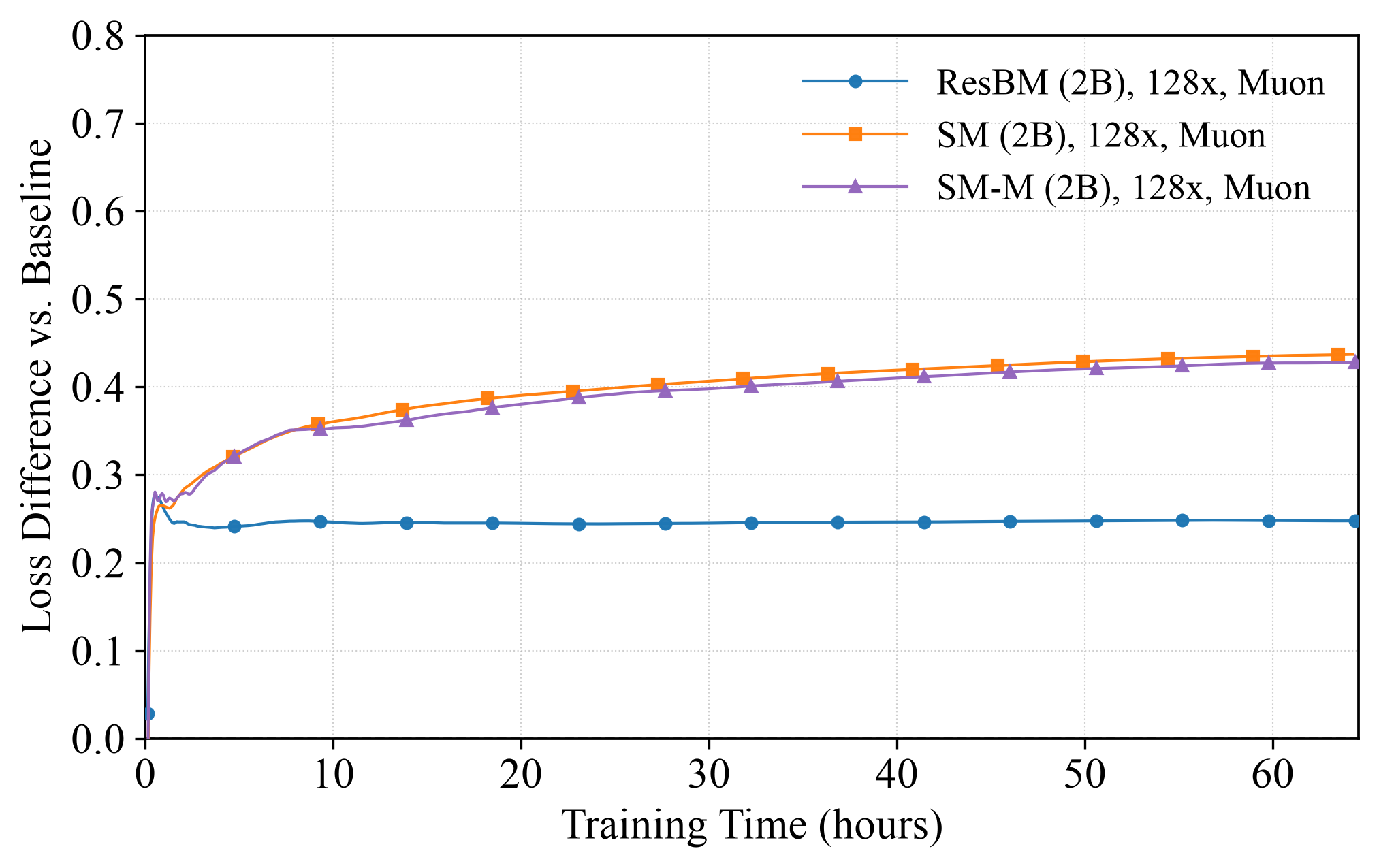}
            \caption{\small Evolution of loss gap (128$\times$).}
            \label{fig:diff_128x_muon}
        \end{subfigure}

    \end{minipage}

    \caption{
        \textbf{Training loss under extreme compression (Muon).} ResBM, SM, SM-M, and SWARM (2B parameters, 4.5--5B tokens) at $100\times$ (a,c) and $128\times$ (b,d) compression; SWARM is shown in (a) only. (c,d) Loss gap to the uncompressed Muon baseline: ResBM's stays flat, while SM's grows.
    }

    \label{fig:sm_vs_resbm_muon}
\end{figure*}

\begin{figure*}[t]
    \centering
    \begin{minipage}[c]{0.58\linewidth}
        \centering

        \begin{subfigure}[t]{\linewidth}
            \centering
            \includegraphics[width=\linewidth]{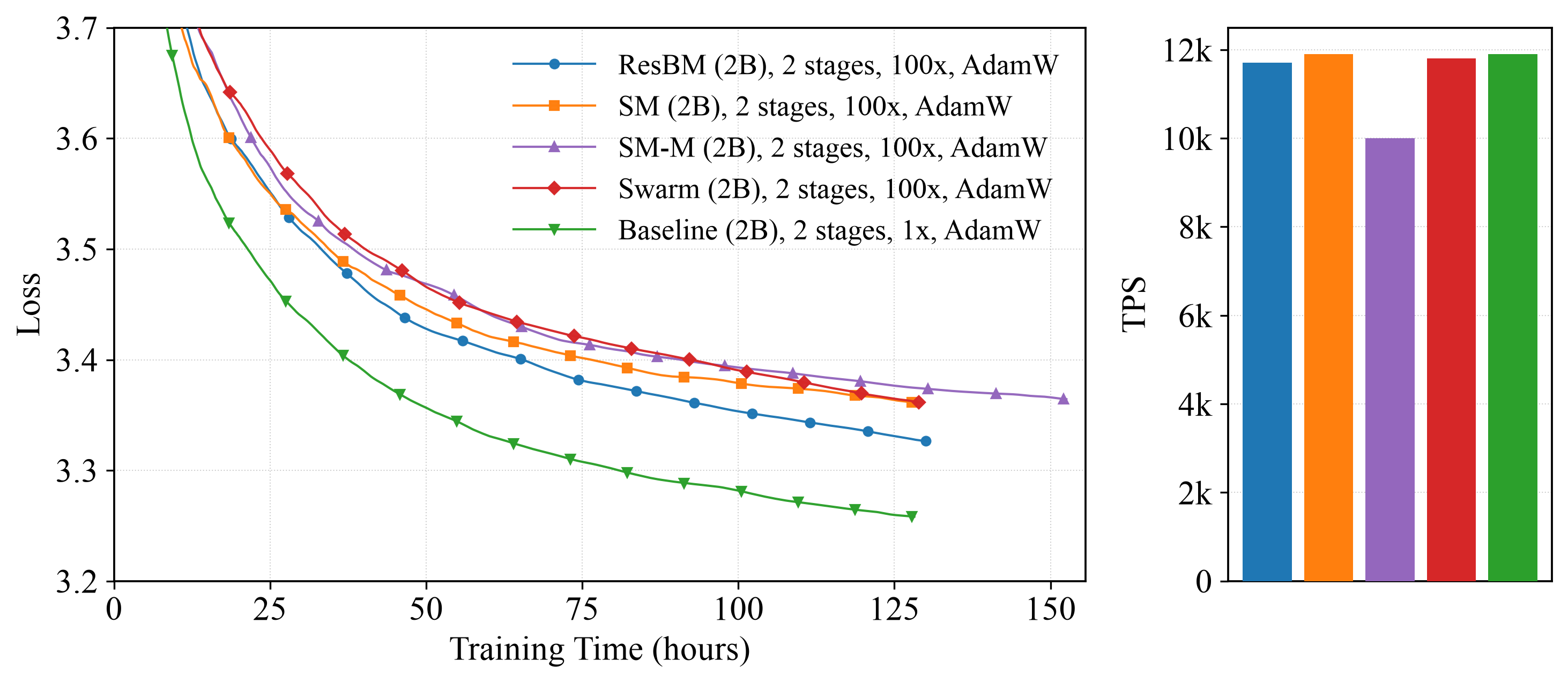}
            \caption{\small Training loss and throughput (2 stages).}
            \label{fig:pp2_loss}
        \end{subfigure}

        \par\smallskip

        \begin{subfigure}[t]{\linewidth}
            \centering
            \includegraphics[width=\linewidth]{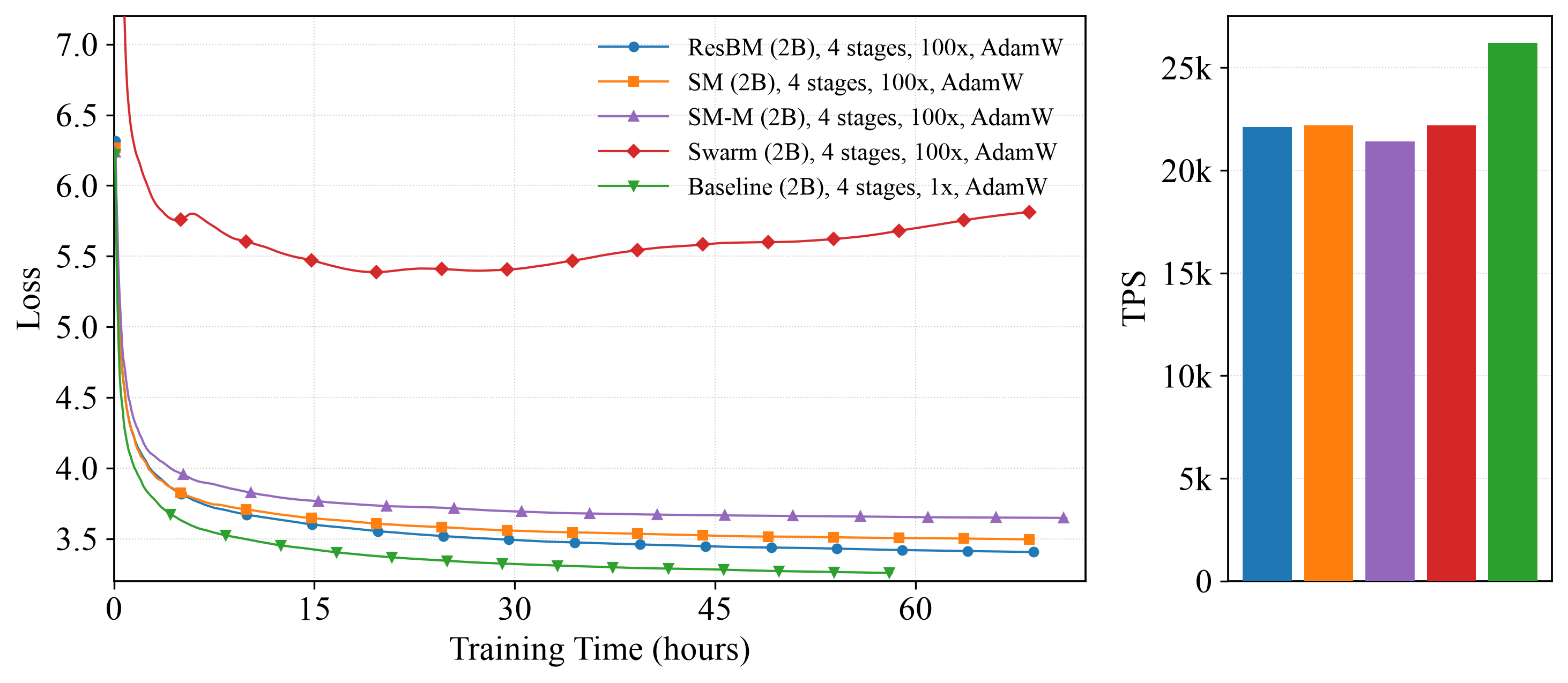}
            \caption{\small Training loss and throughput (4 stages).}
            \label{fig:pp4_loss}
        \end{subfigure}

    \end{minipage}
    \hfill
    \begin{minipage}[c]{0.41\linewidth}
        \centering

        \begin{subfigure}[t]{\linewidth}
            \centering
            \includegraphics[width=\linewidth]{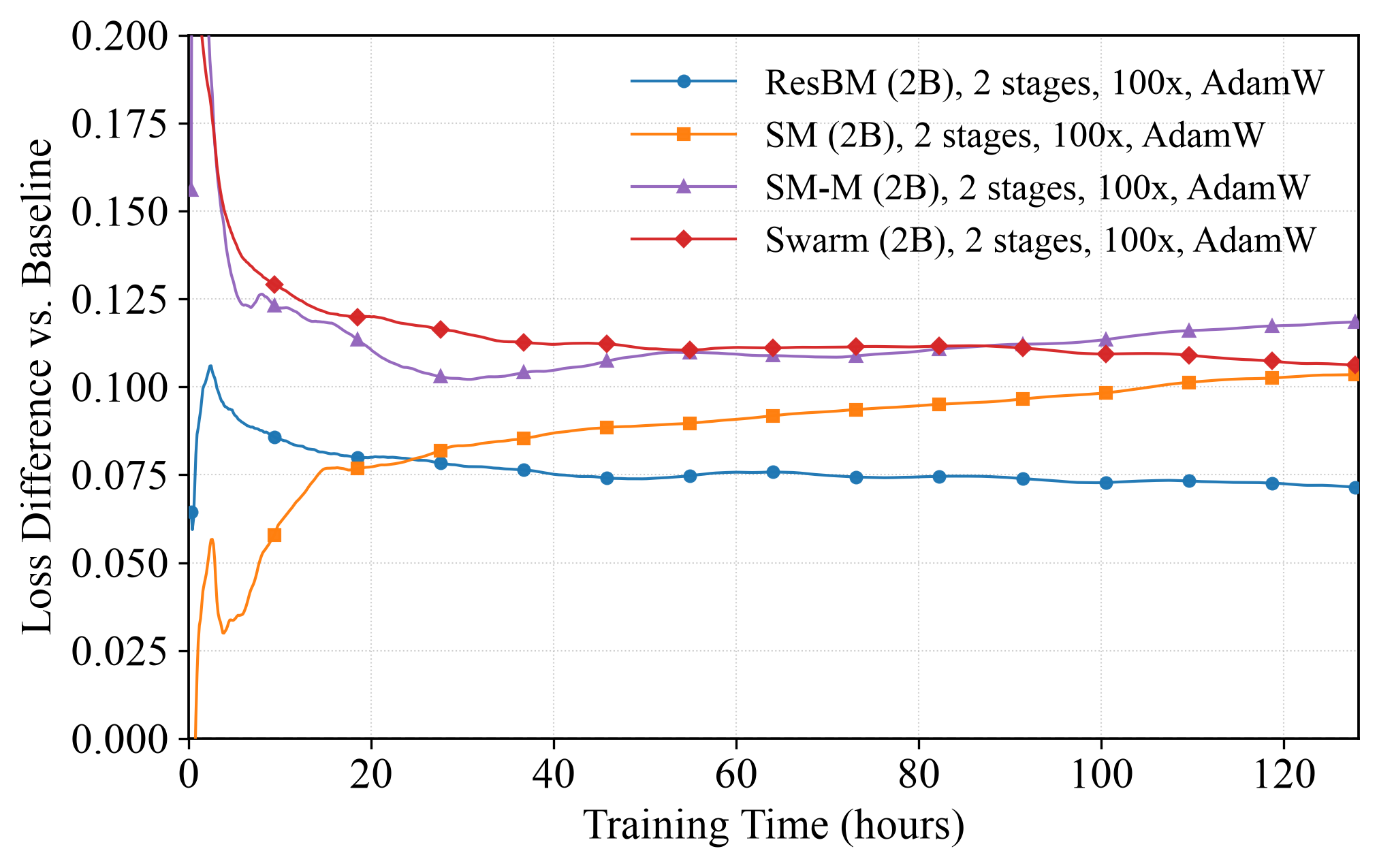}
            \caption{\small Evolution of loss gap (2 stages).}
            \label{fig:pp2_gap}
        \end{subfigure}

        \par\smallskip

        \begin{subfigure}[t]{\linewidth}
            \centering
            \includegraphics[width=\linewidth]{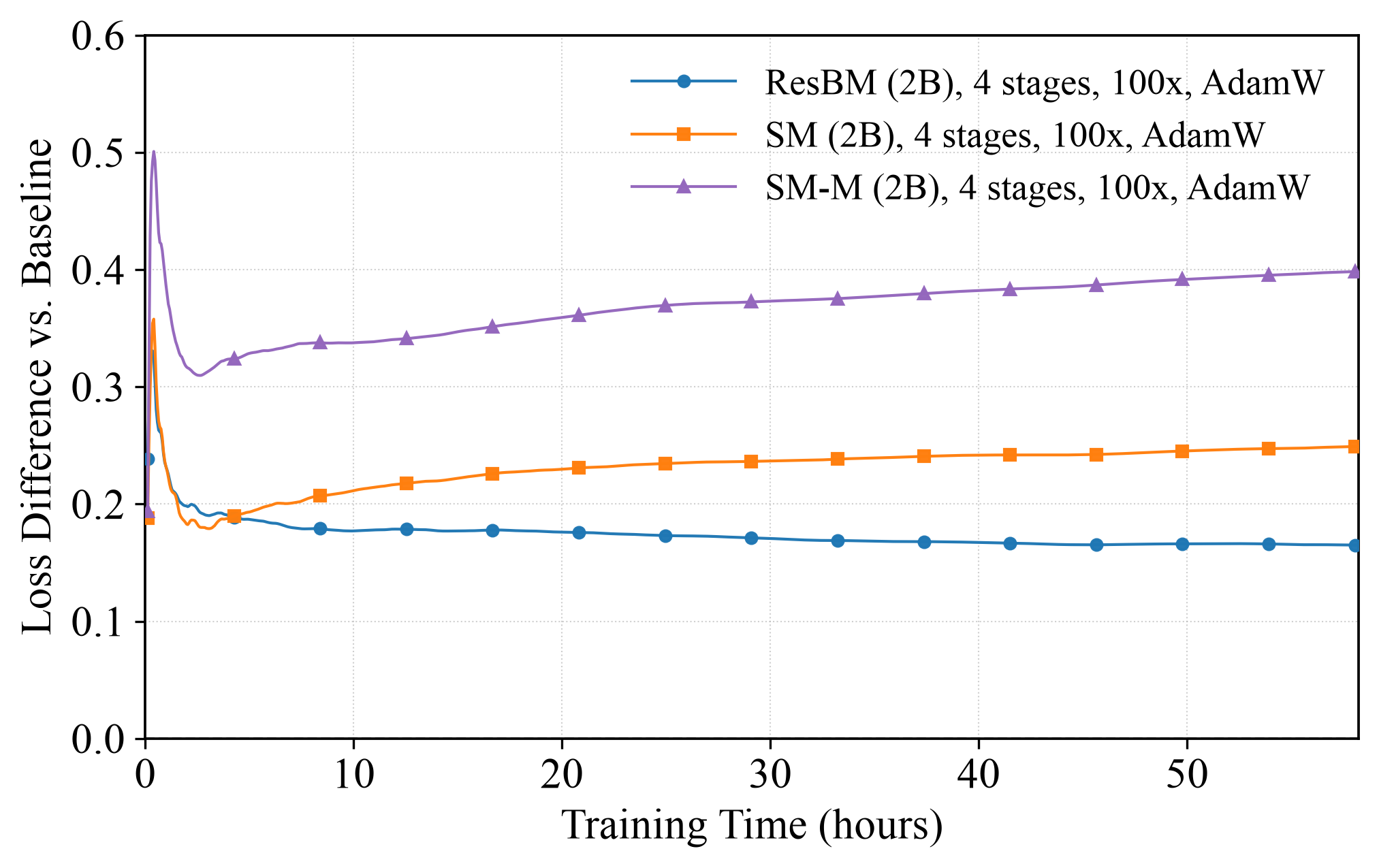}
            \caption{\small Evolution of loss gap (4 stages).}
            \label{fig:pp4_gap}
        \end{subfigure}

    \end{minipage}
    \caption{
        \textbf{Training loss with fewer pipeline stages (AdamW).} ResBM, SM, SM-M, and SWARM (2B parameters, about 5.5B tokens, $100\times$ compression) with 2 (a,c) and 4 (b,d) pipeline stages. (c,d) Loss gap to the uncompressed AdamW baseline: ResBM's stays flat, while SM's grows; SWARM is omitted from (d) because its gap lies far outside the plotted range. ResBM with 2, 4, and 8 stages is compared in Appendix Figure~\ref{fig:resbm_pp_stages_loss}.
    }
    \label{fig:varying_split_count}
\end{figure*}

Each model is trained on about 5B tokens under the configuration of Section~\ref{sec:experiment_setup}, alongside an uncompressed baseline trained with the same optimizer (AdamW).
As shown in Figure~\ref{fig:resbm_vs_all_high_comp}, ResBM outperforms SM and SM-M at both compression rates.
ResBM's loss gap to the uncompressed baseline remains approximately stable over training, whereas the gaps of SM and SM-M grow.
Although SM-M trains more parameters than SM, the two converge almost identically.
We attribute this to where the additional parameters are placed: every block SM-M adds lies before the last pipeline boundary, so its output projection is restricted to the shared low-dimensional subspace (13 of SM-M's 14 blocks, against 7 of SM's 8; Appendix~\ref{app:sm_m}, Table~\ref{tab:sm_m_arch}), which likely limits the benefit of the extra capacity.
SWARM does not train in this regime, its loss staying near its initial value (Appendix Figure~\ref{fig:swarm_all_loss}), so we omit it from Figure~\ref{fig:resbm_vs_all_high_comp}. As a sanity check, the same figure shows its training improving progressively as the number of pipeline stages is reduced to 4, 3, and 2.
Throughput follows a consistent pattern across all our experiments (Figures~\ref{fig:resbm_vs_all_high_comp}--\ref{fig:isostage}). With the same backbone, ResBM, SM, and SWARM reach nearly the same throughput, within about 2\%, SM being marginally faster as it trains fewer parameters; SM-M is slower because of its additional transformer blocks.
Training is mostly bound by activation transfer. Since the baseline's 10~Gbps link is $125\times$ faster than the 80~Mbps link of the compressed models, a compressed transfer still takes longer than the baseline's at $100\times$ compression, but not at $128\times$. Accordingly, at $100\times$ the compressed models reach about 75--85\% of the baseline throughput with four or more pipeline stages, whereas at $128\times$ ResBM and SM match it.

\paragraph{Optimizing with Muon.}
To evaluate ResBM under a different optimizer, we replace AdamW with Muon~\citep{jordan2024muon}, using a Nesterov momentum of $0.95$ and a peak learning rate of $5\times10^{-3}$ with the same schedule, at both compression rates. As shown in Figure~\ref{fig:sm_vs_resbm_muon}, ResBM reaches a lower loss than SM, SM-M, and SWARM under Muon optimization. As with AdamW, SM-M brings no improvement over SM despite its additional parameters. More importantly, switching from AdamW to Muon reduces ResBM's loss gap to the uncompressed baseline by roughly $0.03$ and $0.04$ nats at $100\times$ and $128\times$, respectively, whereas the gaps of SM and SM-M remain approximately unchanged.

We hypothesize that this difference is related to how the two methods use the available representation space. SM constrains most of the model to a shared low-rank subspace, whereas ResBM leaves the main representation unconstrained and learns a dedicated encoder from the full activation stream. Since Muon exhibits less rank collapse than AdamW (Appendix Figure~\ref{fig:svd_analysis}), this may benefit ResBM more directly by allowing its bottleneck to exploit a richer representation before compression.

We also find that SWARM trains under Muon but remains far behind the other methods, though less dramatically than under AdamW. This suggests that SWARM is more prone to training instability, possibly because its bottlenecks interrupt the identity path (Section~\ref{sec:identity_path}); more thorough hyperparameter tuning, which we did not perform for any method, may narrow this gap and is left for future work.

\paragraph{Varying the number of pipeline stages.}
To test whether ResBM's advantage is specific to the extreme setting in which compression is applied after every transformer block, we keep the compression rate at $100\times$ and reduce the number of pipeline stages to 2 and 4. For ResBM, SWARM, and SM, this places a single boundary after the fourth block with 2 stages, and boundaries after the first, fourth, and seventh blocks with 4 stages. SM-M, which has more blocks, places its boundary after its eighth block with 2 stages, and after its fourth, eighth, and twelfth blocks with 4 stages (Tables~\ref{tab:sm_m_arch_pp2} and~\ref{tab:sm_m_arch_pp4}). As shown in Figure~\ref{fig:varying_split_count}, ResBM retains the lowest loss in both settings, and its loss gap to the uncompressed baseline again stays approximately stable, whereas SM's grows. SM-M again brings no improvement over SM: with 2 stages the two reach about the same loss for the same number of tokens, and with 4 stages SM-M underperforms SM by about $0.15$ nats despite holding five more subspace-constrained blocks (12 against 7). With 2 stages, SWARM is on par with SM, but with 4 its loss gap is substantially larger than those of the other methods, so we omit it from Figure~\ref{fig:pp4_gap}.

\paragraph{Scaling model size and context length.}
We further scale the model to 3.5B parameters split into 16 pipeline stages, and, for the 2B model, extend the context length to 2048 tokens on a different dataset, FineWeb-Edu2~\citep{penedo2024finewebdatasetsdecantingweb}. In both settings, ResBM keeps its lead over SM with the same dynamics: its loss gap to the uncompressed baseline stays approximately stable, whereas SM's grows (Appendix~\ref{app:scaling_loss}).

\subsection{ResBM in practice}
\label{sec:practical_considerations}

Taken together, our experiments show that ResBM consistently outperforms previous compression methods across different regimes. Beyond this overall advantage, three practical observations emerge.

First, after the initial phase of training, ResBM exhibits an approximately constant loss offset relative to the uncompressed baseline. This suggests ResBM may preserve the token-scaling behavior of the underlying model, which could make existing scaling laws useful for predicting compressed training performance.

\begin{figure*}[t]
    \centering
    \begin{subfigure}[t]{0.58\linewidth}
        \centering
        \includegraphics[width=\linewidth]{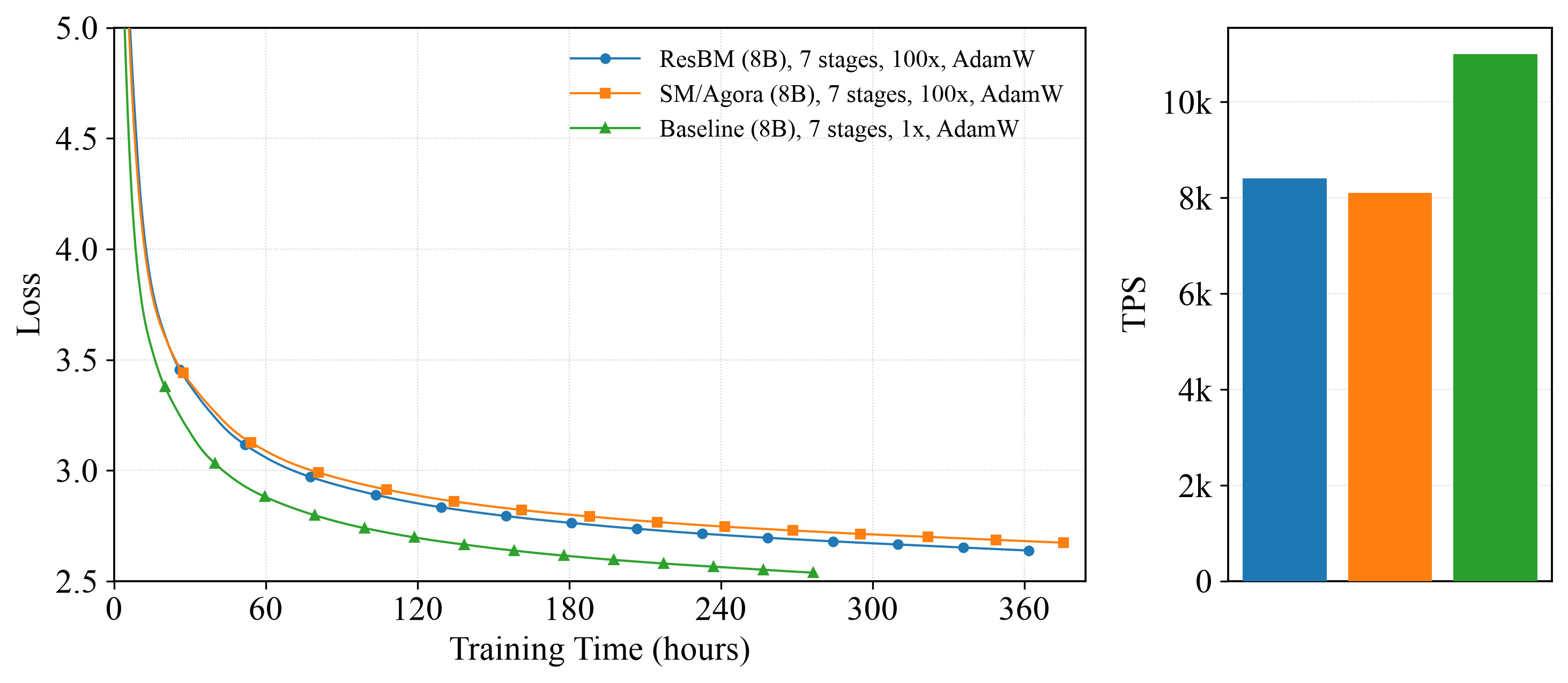}
        \caption{\small Training loss and throughput (8B).}
        \label{fig:isostage_loss}
    \end{subfigure}
    \hfill
    \begin{subfigure}[t]{0.41\linewidth}
        \centering
        \includegraphics[width=\linewidth]{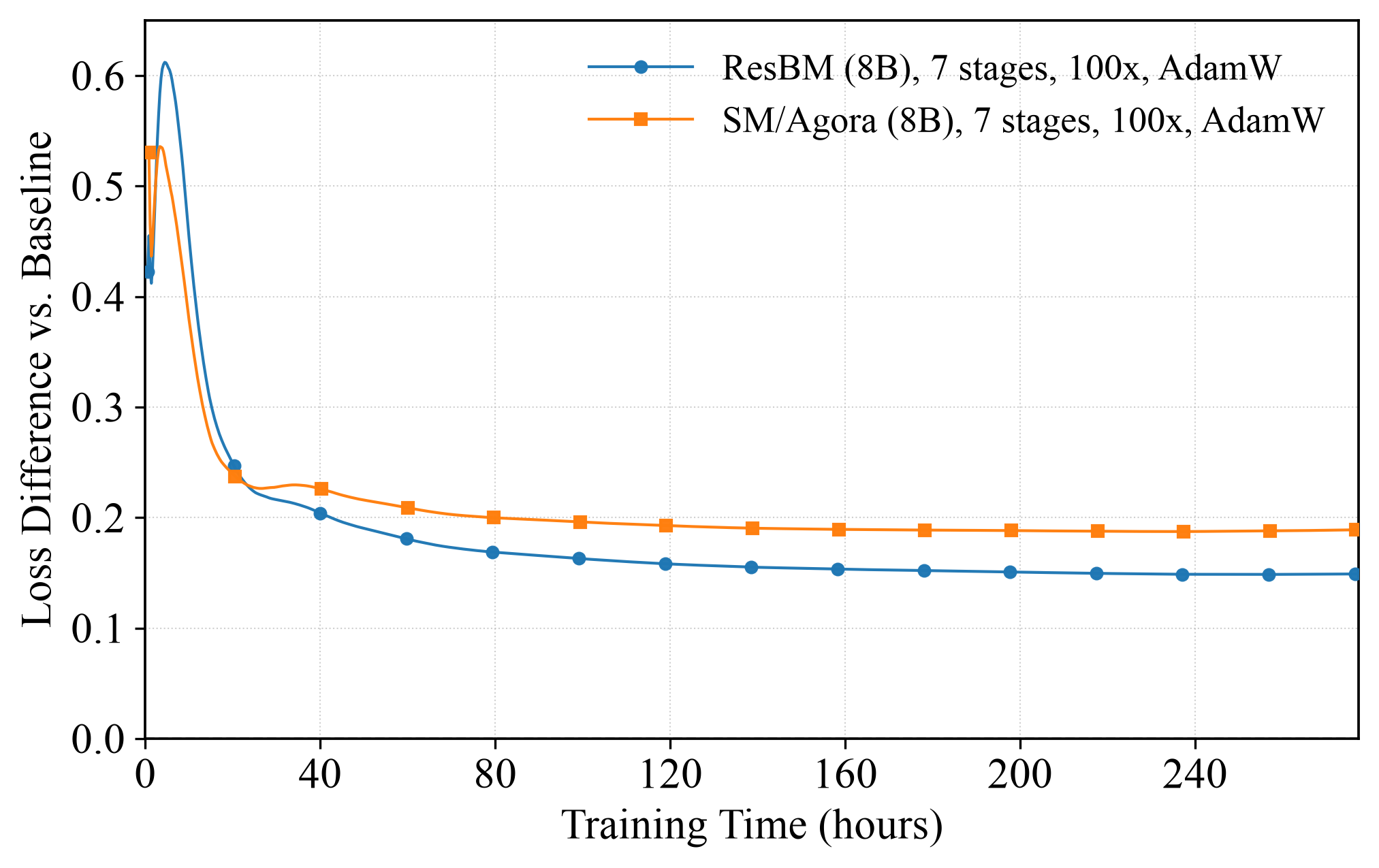}
        \caption{\small Evolution of loss gap (8B).}
        \label{fig:isostage_gap}
    \end{subfigure}
    \caption{
        \textbf{Training loss at 8B scale under a practical pipeline layout.} ResBM and SM on the seven-stage layout of the public Agora-8B run (roughly $100\times$ compression, 11B tokens). (b) Loss gap to the uncompressed 8B baseline. Details in Appendix~\ref{app:isostage}.
    }
    \label{fig:isostage}
\end{figure*}

Second, our main experiments deliberately apply compression after almost every transformer block, whereas in practice the number of pipeline stages and the blocks assigned to each depend on the memory of the participating nodes. Reducing the number of stages preserves ResBM's advantage while bringing its loss much closer to the uncompressed baseline (Figure~\ref{fig:varying_split_count}), suggesting that practical deployments may incur substantially smaller degradation than our most extreme settings. To test such a layout at a larger scale, we reproduce Agora-8B, the public 8B-parameter SM run of \citet{avraham2026agora}, whose seven pipeline stages are sized to consumer and workstation GPUs, and match ResBM to it stage by stage in trainable parameters at roughly $100\times$ compression (Appendix~\ref{app:isostage}). Trained for 11B tokens, ResBM again outperforms SM, with a smaller loss gap to the uncompressed baseline: about $0.15$ nats by the end of training, against $0.19$ for SM (Figure~\ref{fig:isostage}). At this budget, it remains unclear whether SM's gap grows as in our 2B experiments; longer training runs are needed to settle this.

\begin{figure}[t]
    \centering
    \begin{subfigure}[t]{0.49\linewidth}
        \centering
        \includegraphics[width=\linewidth]{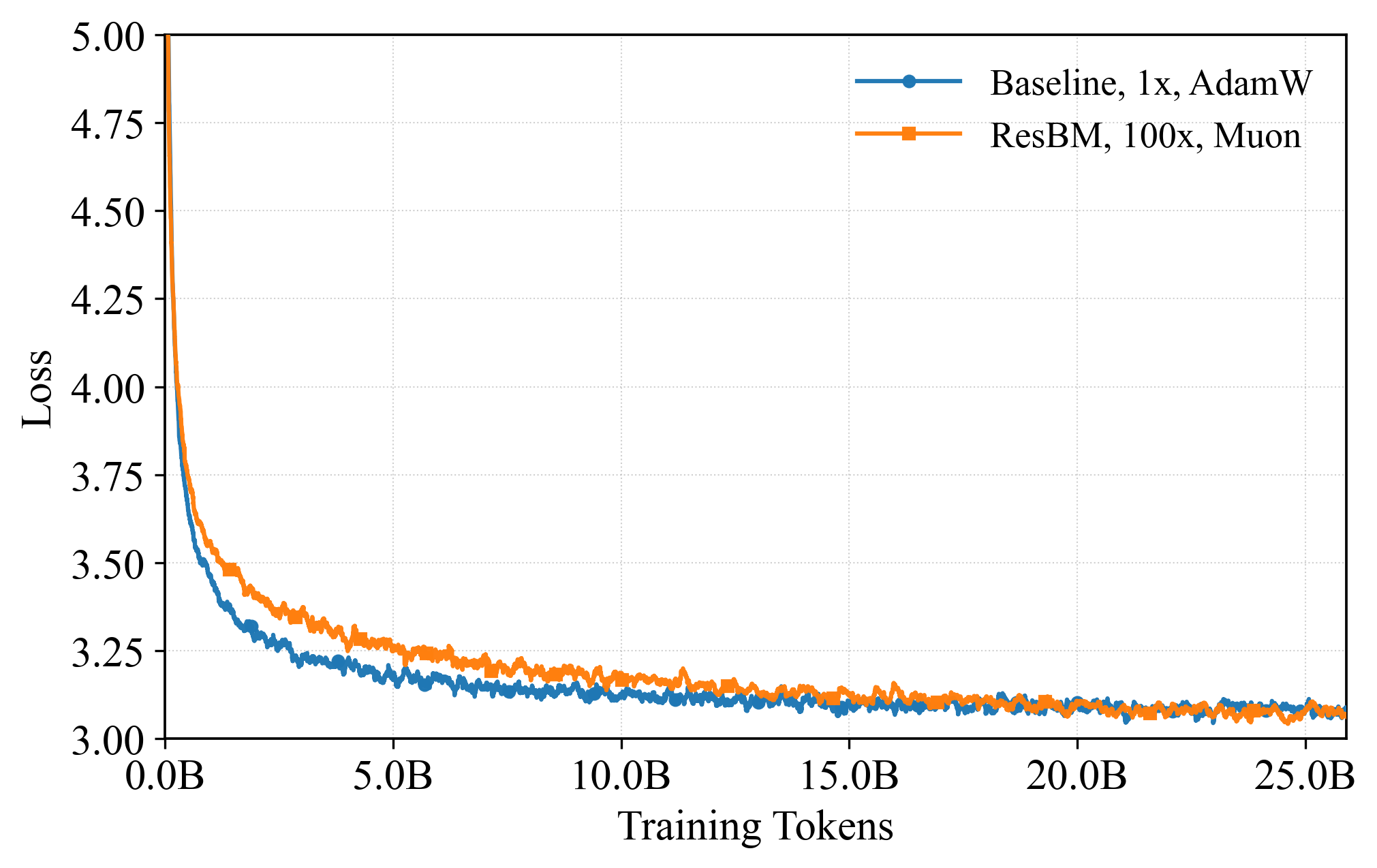}
        \caption{\small Training loss ($100\times$).}
        \label{fig:muon_vs_adamw_100x}
    \end{subfigure}\hfill
    \begin{subfigure}[t]{0.49\linewidth}
        \centering
        \includegraphics[width=\linewidth]{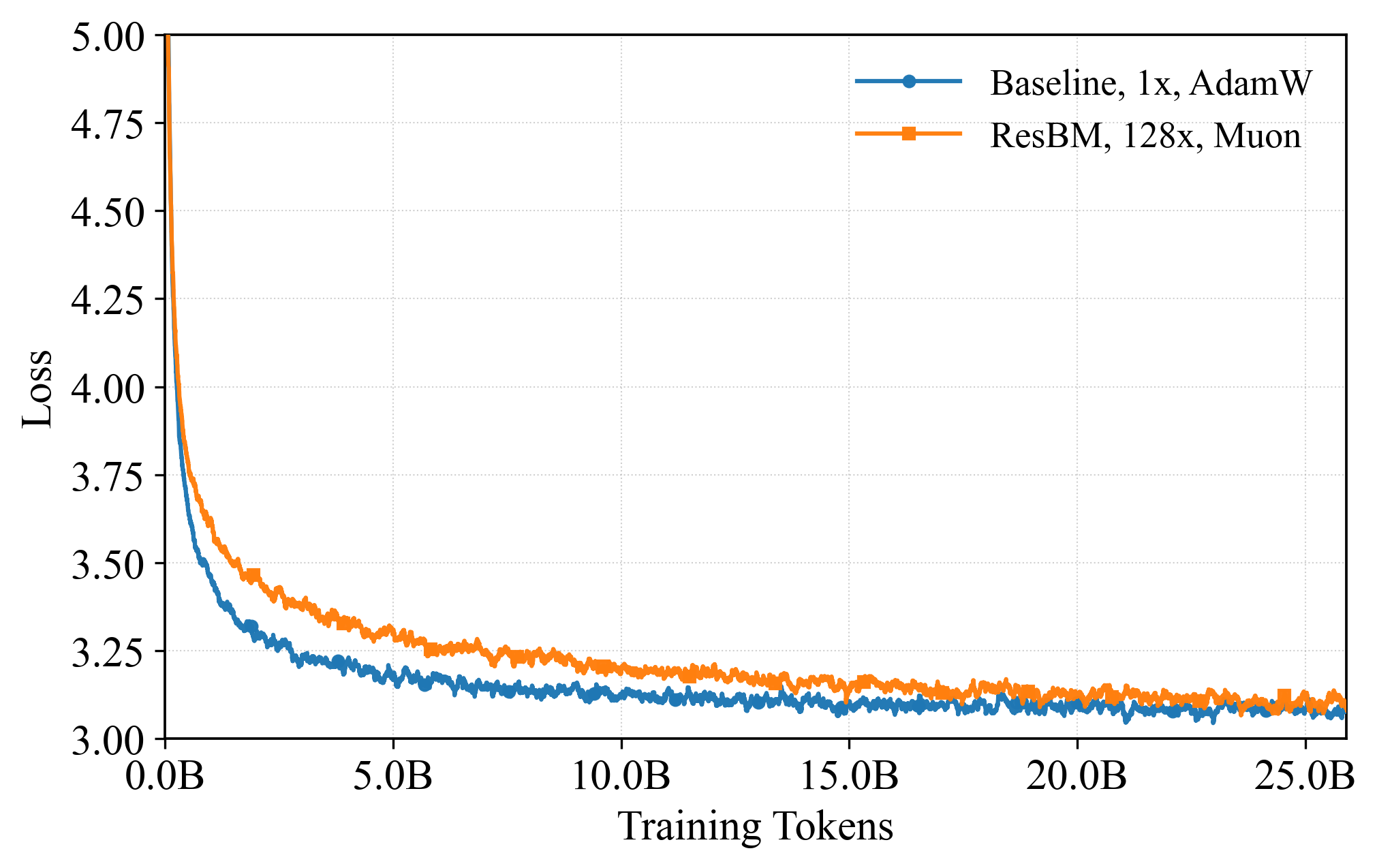}
        \caption{\small Training loss ($128\times$).}
        \label{fig:muon_vs_adamw_128x}
    \end{subfigure}
    \caption{
        \textbf{ResBM with Muon against the uncompressed AdamW baseline.} Training loss of 2B models over 26B tokens at $100\times$ (a) and $128\times$ (b) compression.
    }
    \label{fig:resbm_muon_vs_adamw}
\end{figure}

Finally, Muon provides a particularly useful practical configuration for ResBM. At $100\times$ and $128\times$ compression, ResBM trained with Muon reaches the loss of the uncompressed AdamW baseline after approximately $17$B and $26$B tokens, respectively (Figure~\ref{fig:resbm_muon_vs_adamw}). Paired with a stronger optimizer, extreme activation compression can therefore match the loss of conventional AdamW training. A gap remains relative to the uncompressed Muon baseline (Figure~\ref{fig:sm_vs_resbm_muon}), however, leaving room for further work on closing the optimizer-matched gap.

\section{Conclusion}
\label{sec:conclusion}

We introduced the Residual Bottleneck Model (ResBM), an architecture designed natively for low-bandwidth pipeline-parallel training. In contrast to SWARM, ResBM retains a partial identity path that bypasses each encoder and each decoder, which our ablation shows stabilizes training. In contrast to SM, and like SWARM, it confines compression to pipeline boundaries, so all of its transformer blocks operate at full rank. These benefits come at a small parameter and compute overhead, and ResBM remains effective at extreme compression rates of $100\times$ and $128\times$.
Our extended pretraining runs show that extreme activation compression incurs a training-loss penalty for every method we evaluate, including our own. ResBM exhibits the smallest late-training gap, which remains approximately stable after the initial phase, whereas SM's gap grows over training in our 2B and 3.5B experiments. On a reproduction of a public 8B-parameter SM run, whose seven pipeline stages are sized to consumer and workstation GPUs, ResBM again achieves a lower loss than SM.
Over an emulated 80~Mbps link with 26~ms latency, ResBM reaches roughly 75--100\% of the throughput that the uncompressed baseline achieves over a 10~Gbps link, and matches it at $128\times$ compression. Taken together, these results establish ResBM as a new state of the art for extreme activation compression in low-bandwidth pipeline-parallel training.

\paragraph{Broader impact.}
By achieving the smallest training-loss penalty among the evaluated methods at $100\times$ and $128\times$ compression, ResBM further reduces the communication barrier to pipeline-parallel pretraining over low-bandwidth links, making decentralized pretraining of large models more practical.
This is a step toward more accessible and, potentially, more environmentally sustainable training, by enabling better utilization of geographically distributed idle compute instead of concentrating capacity in a small number of hyperscale datacenters.\footnote{\ifanon ResBM already powers a real-world decentralized pretraining project, demonstrating its applicability beyond controlled experimental settings. The code to reproduce the proposed architecture is open source and available as part of that project's codebase.\else ResBM already powers a real-world decentralized pretraining project named IOTA demonstrating its applicability beyond controlled experimental settings. The code to reproduce the proposed architecture is available as part of the IOTA codebase: \url{https://github.com/macrocosm-os/iota/blob/1de522ea35f7f1fd4f1374e96e2681c51a0739ae/shared/subnet/src/subnet/model/llama3/full.py\#L174}.\fi}

\paragraph{Limitations.}
Our experiments cover 2B, 3.5B, and 8B parameter models, main comparisons up to 11B tokens with supplementary 2B runs up to 26B tokens, C4 and FineWeb-Edu2, context lengths of 1024 and 2048, AdamW and Muon, compression rates of $100\times$ and $128\times$, and 2, 4, 7, 8, and 16 pipeline stages. Nevertheless, frontier-scale models, substantially longer training, additional architecture families, and downstream tasks remain unexplored. Due to compute constraints, no large-scale hyperparameter sweeps were performed; all methods share settings drawn from common practice, prior work and short small-scale runs.
In our 2B and 3.5B experiments, ResBM's loss gap is front-loaded and approximately stable, whereas SM's grows during training. These dynamics require validation at larger model and token scales.
Further exploration of the bottleneck design space is also left to future work.

\clearpage
\ifanon\else
\subsubsection*{Acknowledgments}

The authors would like to thank \textbf{Brian McCrindle} and \textbf{Felix Quinque} for their insightful feedback on this manuscript and for their ongoing efforts in leading the integration of this architecture into the IOTA production framework with \textbf{Szymon Fonau}, \textbf{Nicholas Miller} and the rest of the IOTA team.
Their work in evaluating ResBM within trustless, decentralized environments is providing invaluable insights that will inform future iterations and advancements of the model.
\fi

\bibliographystyle{iclr2027_conference}
\bibliography{references}

\clearpage
\appendix
\counterwithin{figure}{section}
\counterwithin{table}{section}
\section{ResBM architecture}

\label{app:figures}

\vspace{0.25cm}
\begin{figure}[h]
    \centering
    \begin{minipage}{0.8\linewidth}
        \centering
        \includegraphics[width=\linewidth]{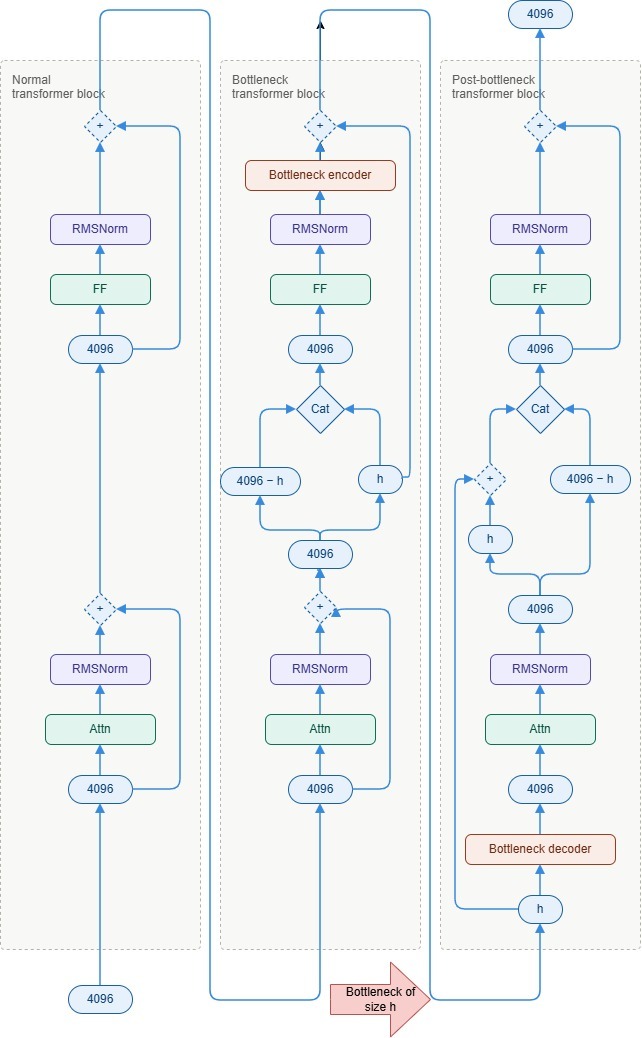}
    \end{minipage}
    \caption{\textbf{ResBM architecture across a pipeline-parallel communication boundary.} For clarity, we show only three transformer blocks and a single encoder--decoder bottleneck instance. The encoder compresses the activation to bottleneck size $h$ before communication, and the decoder expands it back to the full-width representation $H$ at the receiving stage.}
    \label{fig:app:bottleneck_layer}
\end{figure}

\clearpage
\section{Spectral analysis}
\subsection{Baseline output projection spectra}
\label{app:baseline_spectra}

\begin{figure}[h]
    \centering
    \includegraphics[width=0.99\textwidth, trim={0 0 0 58}, clip]{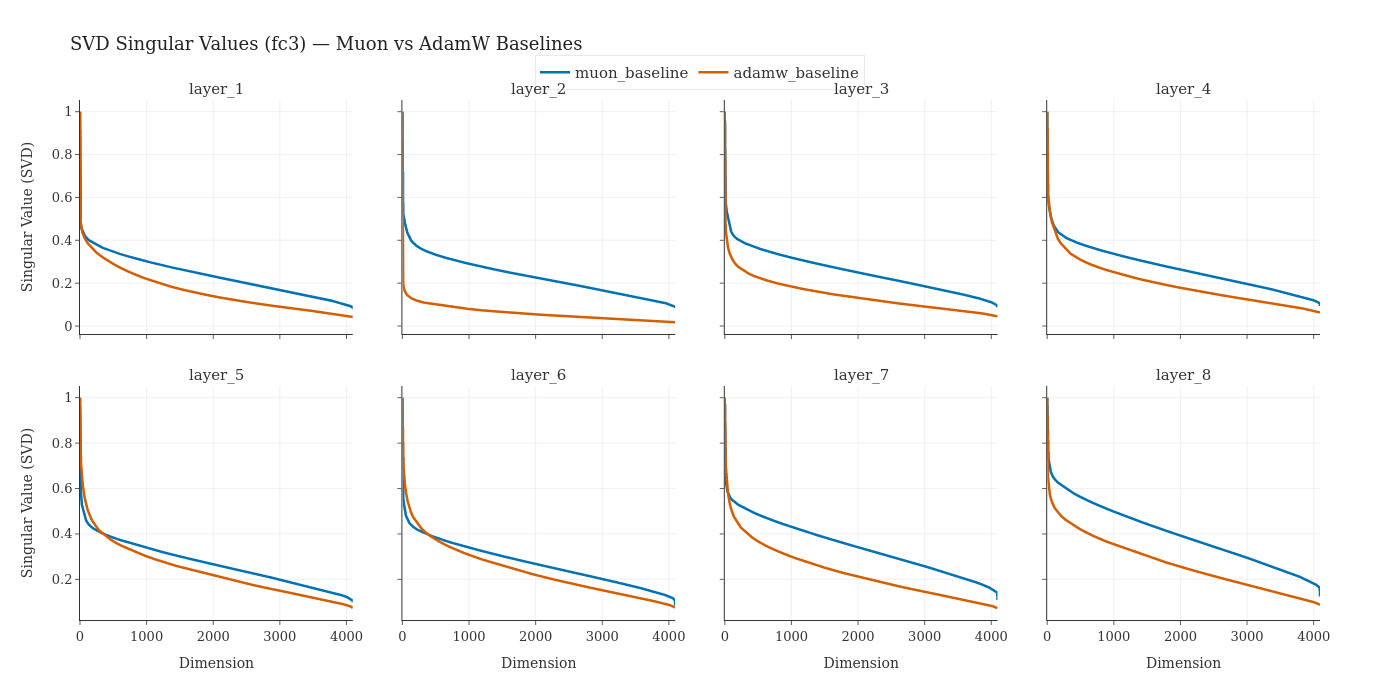}
    \caption{\textbf{Singular value analysis of the FFN output projection matrices of the uncompressed baseline.} Both AdamW and Muon exhibit substantial singular value decay, indicating a significant degree of effective rank collapse. However, AdamW shows a markedly faster decay, whereas Muon maintains larger singular values across more dimensions,
    suggesting weaker effective rank collapse than AdamW.}
    \label{fig:svd_analysis}
\end{figure}

Figure~\ref{fig:svd_analysis} shows the singular value spectra of the FFN output projection matrices of the uncompressed 2B baseline trained with AdamW and with Muon, computed following the protocol of \citet{liu2025muonscalablellmtraining}.

\clearpage
\section{Additional convergence experiments}
\subsection{SWARM loss under different numbers of pipeline stages}
\label{app:swarm_loss}

\begin{figure}[h]
    \centering
    \includegraphics[width=0.8\textwidth]{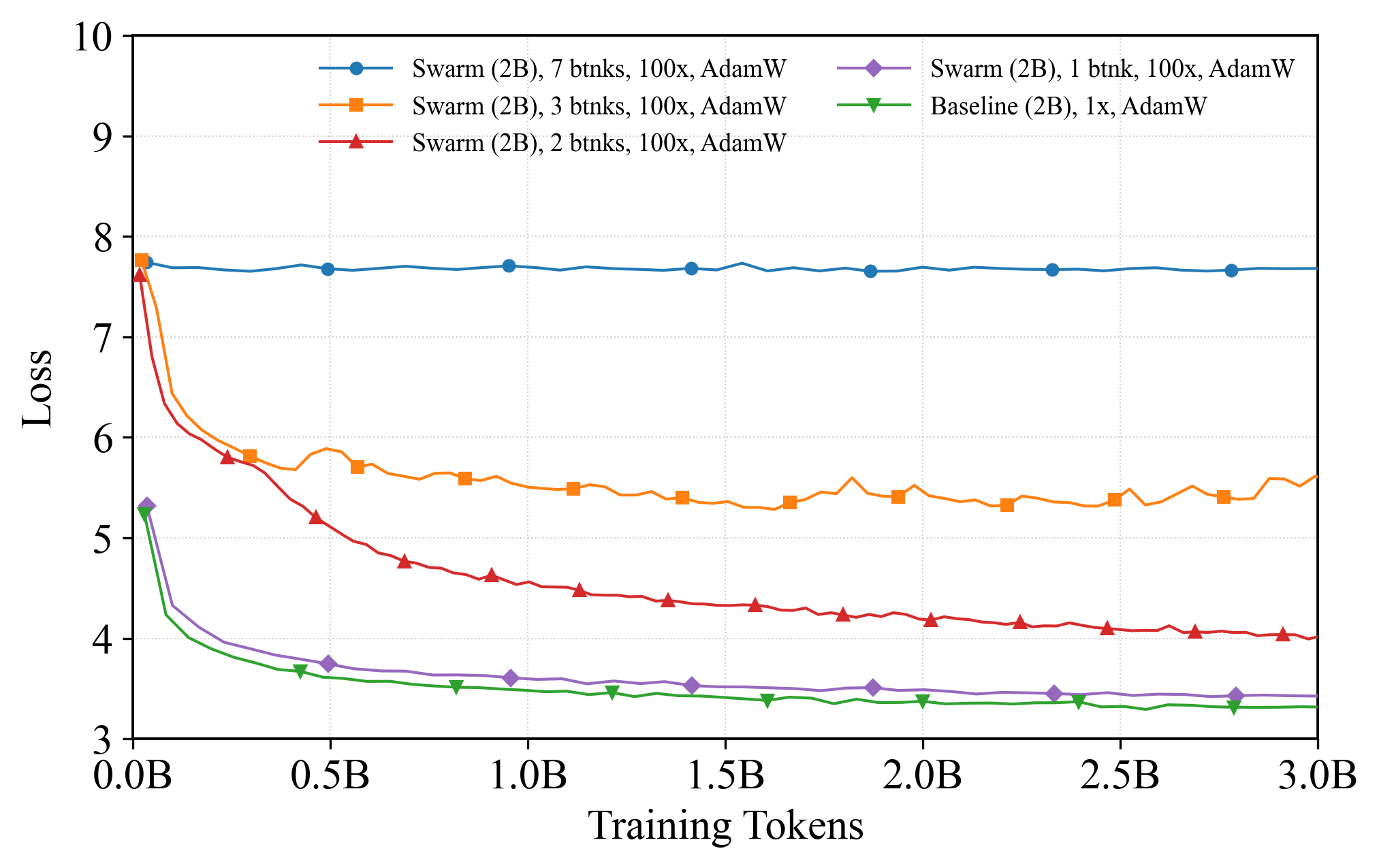}
    \caption{\textbf{SWARM convergence under varying numbers of pipeline stages.} Training loss for SWARM with 2, 3, 4, and 8 pipeline (PP) stages, showing that convergence degrades significantly as the number of PP stages increases. The plot legend counts communication boundaries (``btnk''), one fewer than the number of PP stages.}
    \label{fig:swarm_all_loss}
\end{figure}

\subsection{ResBM loss under different numbers of pipeline stages}
\label{app:resbm_pp_stages}

\begin{figure}[h]
    \centering
    \includegraphics[width=0.8\textwidth]{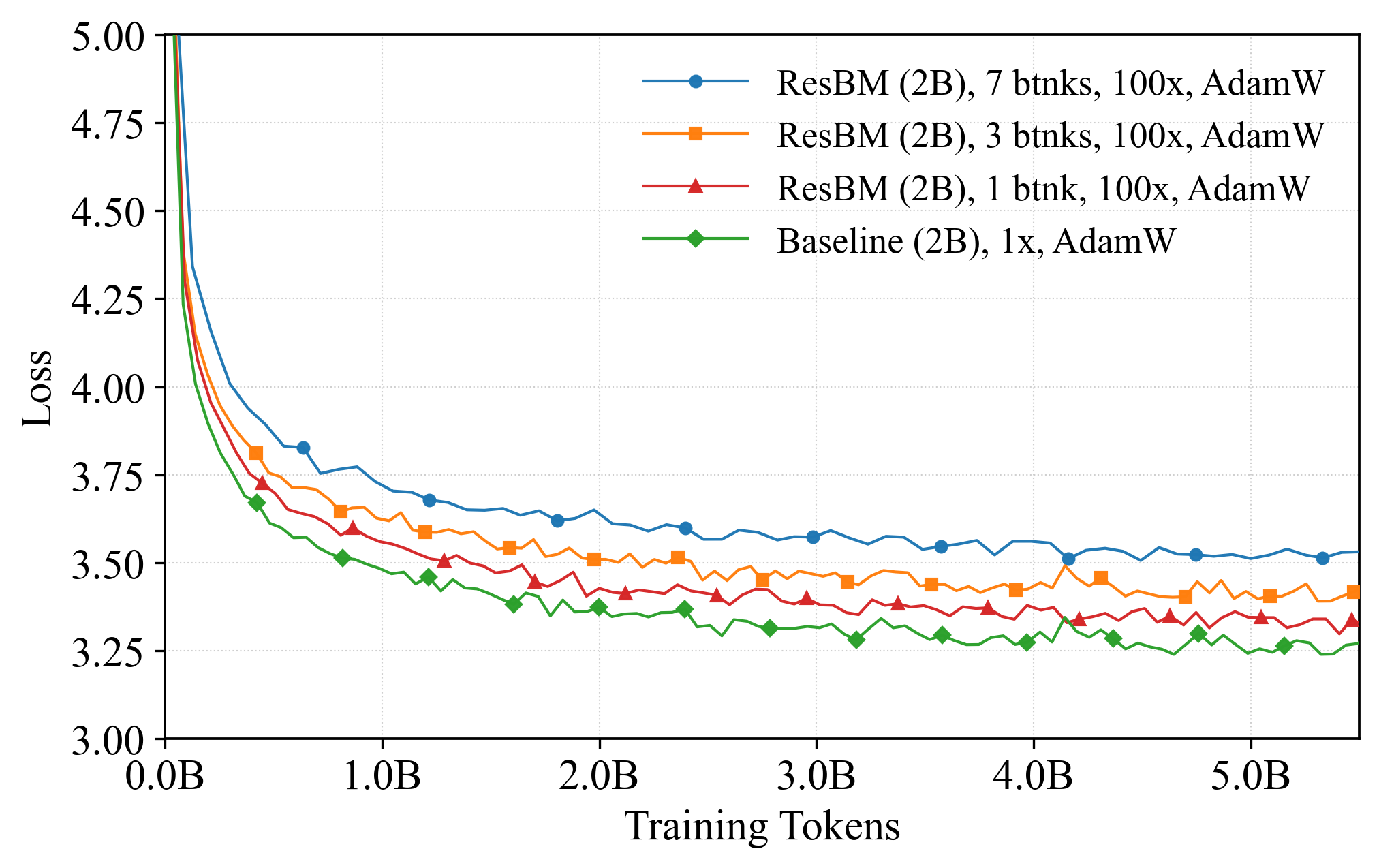}
    \caption{\textbf{ResBM convergence under varying numbers of pipeline stages.} Training loss of 2B-parameter ResBM models trained with AdamW at $100\times$ compression and split into 2, 4, and 8 pipeline (PP) stages, together with the uncompressed AdamW baseline. Each boundary between consecutive stages holds one bottleneck, so these configurations correspond to the 1, 3, and 7 bottlenecks shown in the legend. ResBM's loss increases with the number of PP stages, with the 2-stage model closest to the baseline.}
    \label{fig:resbm_pp_stages_loss}
\end{figure}

\subsection{Loss under model-size and context-length scaling}
\label{app:scaling_loss}

\begin{figure}[h]
    \centering
    \begin{minipage}{0.95\linewidth}
        \centering
        \begin{subfigure}[t]{0.48\linewidth}
            \centering
            \includegraphics[width=\linewidth]{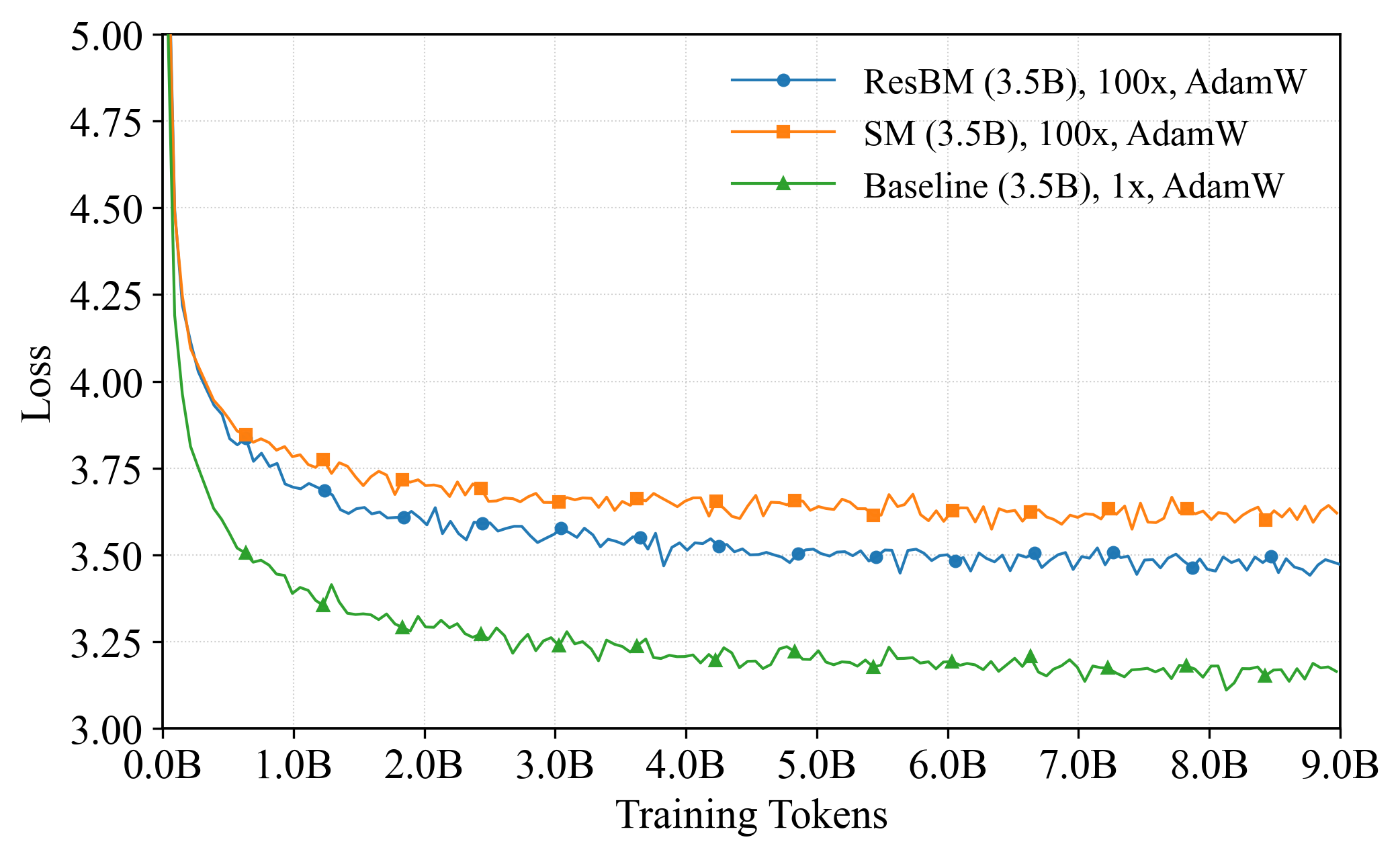}
            \caption{\small Training loss (3.5B, 16 stages).}
            \label{fig:scaling_3-5B_loss}
        \end{subfigure}%
        \hfill
        \begin{subfigure}[t]{0.48\linewidth}
            \centering
            \includegraphics[width=\linewidth]{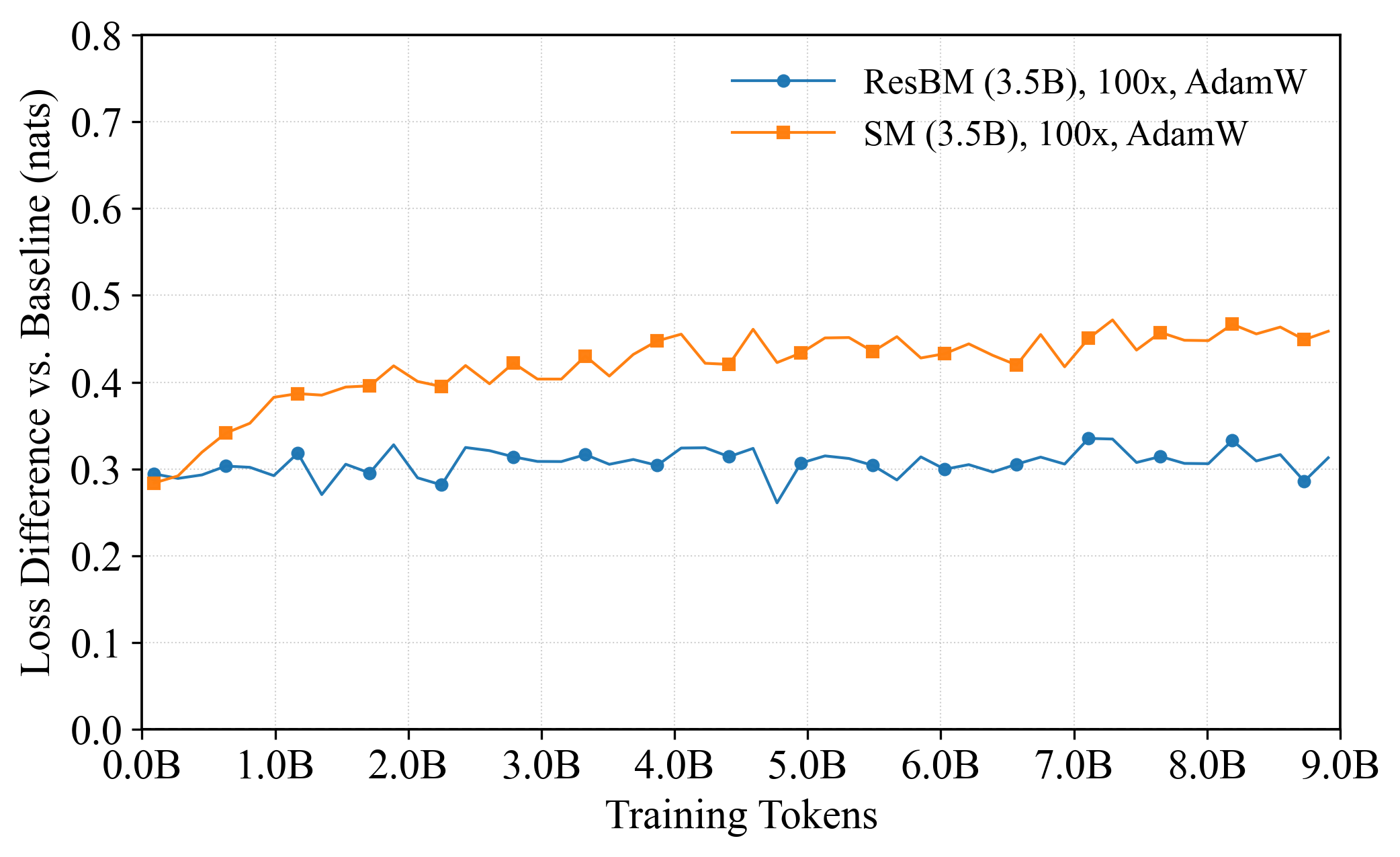}
            \caption{\small Evolution of loss gap (3.5B, 16 stages).}
            \label{fig:scaling_3-5B_gap}
        \end{subfigure}
    \end{minipage}
    \begin{minipage}{0.95\linewidth}
        \centering
        \begin{subfigure}[t]{0.49\linewidth}
            \centering
            \includegraphics[width=\linewidth]{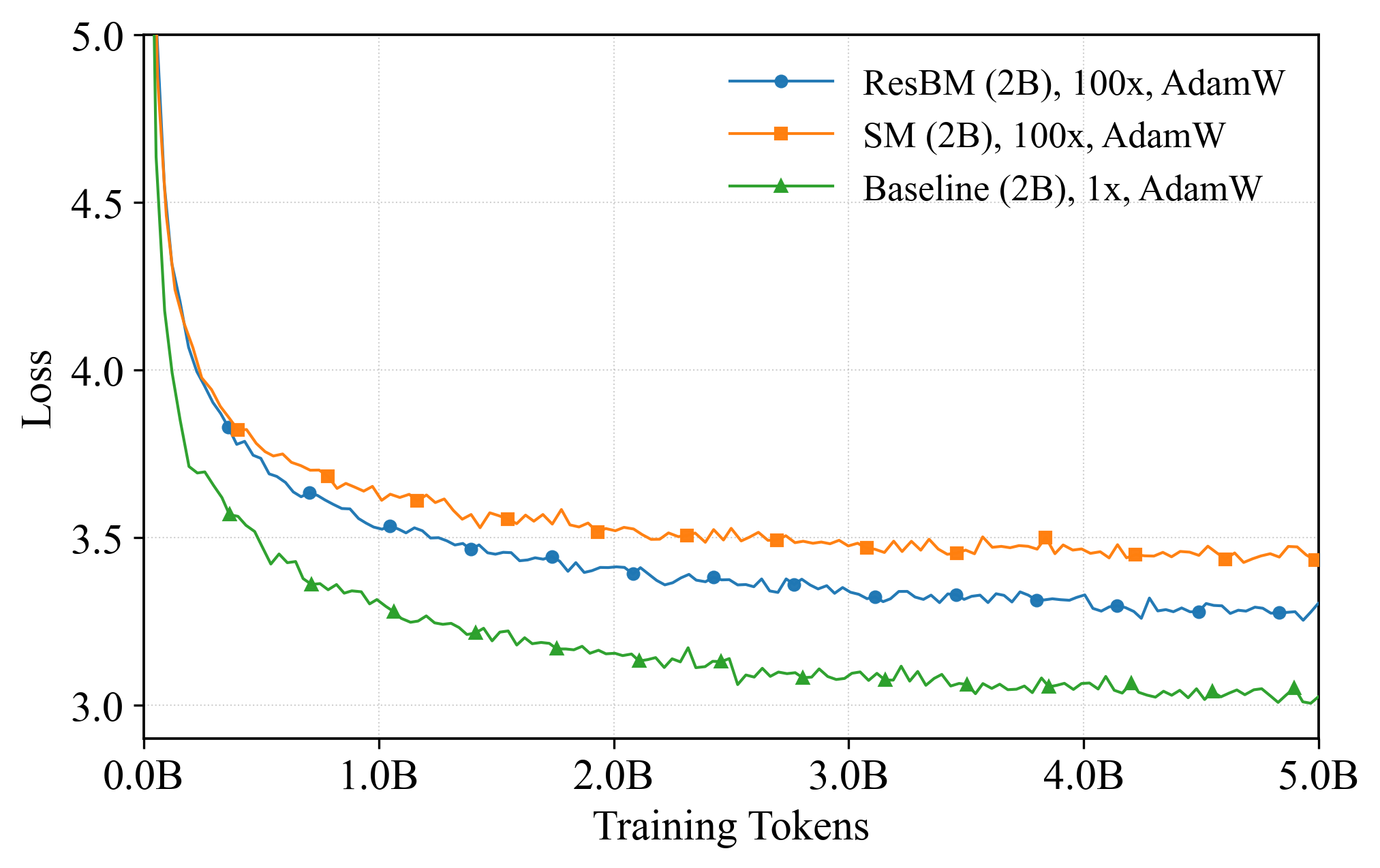}
            \caption{\small Training loss (context length 2048).}
            \label{fig:scaling_ctx2048_loss}
        \end{subfigure}\hfill
        \begin{subfigure}[t]{0.49\linewidth}
            \centering
            \includegraphics[width=\linewidth]{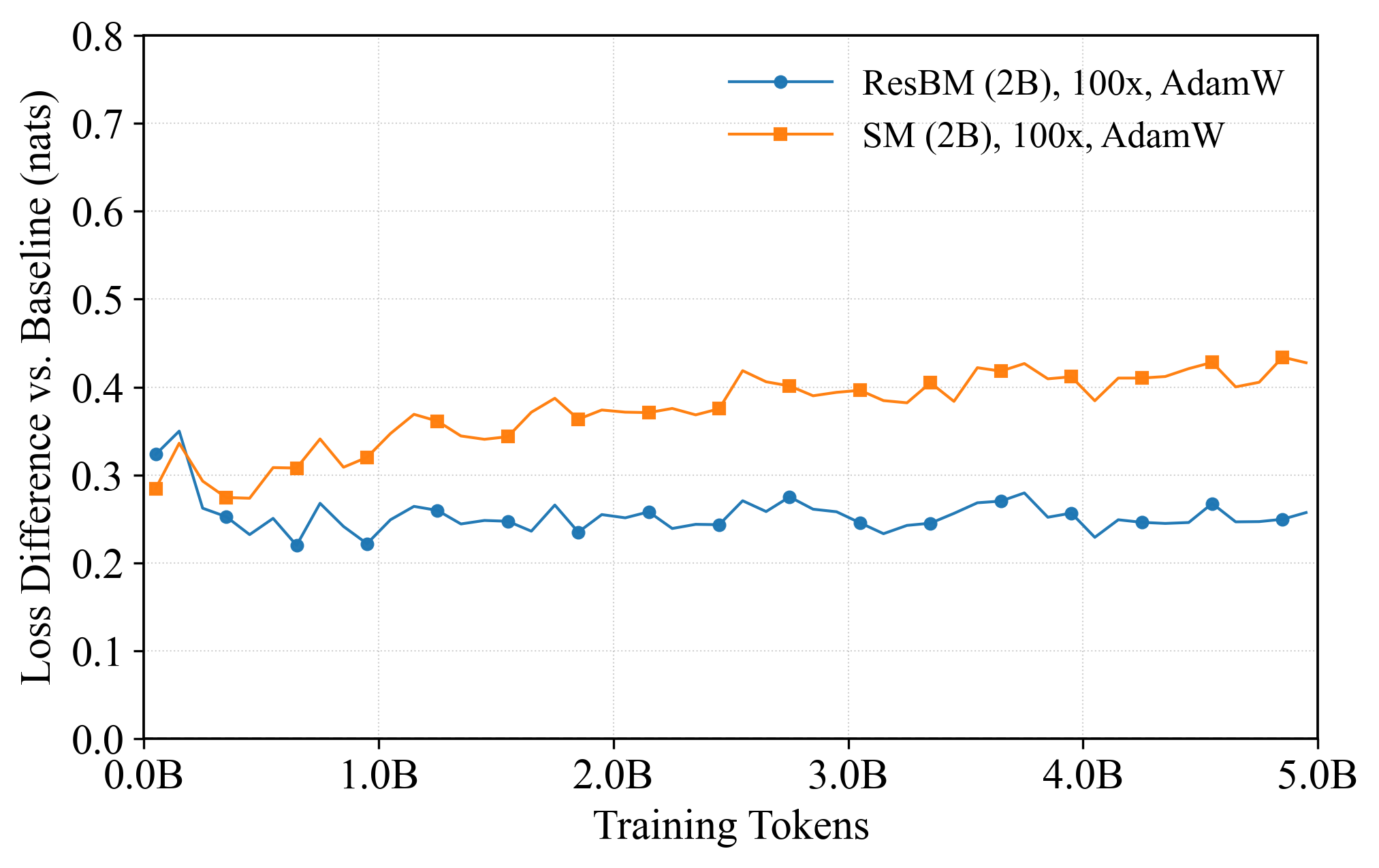}
            \caption{\small Evolution of loss gap (context length 2048).}
            \label{fig:scaling_ctx2048_gap}
        \end{subfigure}
    \end{minipage}
    \caption{\textbf{Loss under model-size and context-length scaling.} ResBM and SM at $100\times$ compression (AdamW): (a,b) a 3.5B model with 16 pipeline stages; (c,d) the 2B model with context length 2048 on FineWeb-Edu2. (b,d) Loss gap to the corresponding uncompressed baseline, at the same number of training tokens.}
    \label{fig:all_vs_resbm_3-5B}
\end{figure}

We evaluate ResBM against SM along two further axes, at $100\times$ compression with AdamW. First, we scale the model to 3.5B parameters with 16 transformer blocks, split into 16 pipeline stages of one block each (15 communication boundaries), and train it for 9B tokens. Second, we extend the context length of the 2B model to 2048 tokens and switch to FineWeb-Edu2 (\texttt{HuggingFaceFW/fineweb-edu-score-2}), which contains curated educational content \citep{penedo2024finewebdatasetsdecantingweb}, and train it for 5B tokens. As shown in Figure~\ref{fig:all_vs_resbm_3-5B}, ResBM achieves a lower loss than SM in both settings. Its loss gap to the uncompressed baseline stays approximately stable over training, at about $0.3$ nats for the 3.5B model and $0.25$ nats with the longer context, whereas SM's grows to about $0.45$ and $0.43$ nats, respectively.

\clearpage
\section{Additional systems experiments}
\label{app:appendix_throughput}

\subsection{Throughput gain as a function of bandwidth}
\label{app:throughput_bandwidth}

\begin{figure}[h]
    \centering
    \includegraphics[width=0.85\linewidth]{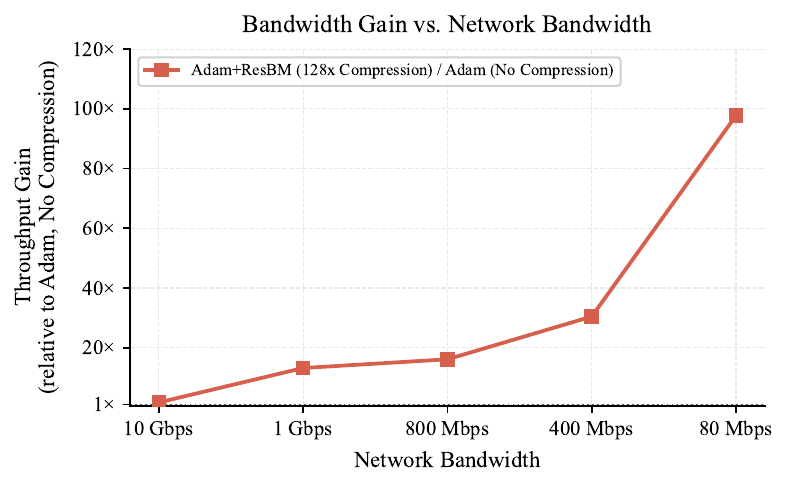}
    \caption{Throughput gain of ResBM ($128\times$ compression) over the uncompressed baseline, both 2B models trained with AdamW, as a function of inter-stage bandwidth.
    The gain is about $1.7\times$ at 10~Gbps and grows as bandwidth drops, reaching about $98\times$ at 80~Mbps.}
    \label{fig:throughput_gain}
\end{figure}

To characterize when activation compression pays off, we measure the training throughput of ResBM ($128\times$) and of the uncompressed baseline, both 2B models trained with AdamW, at inter-stage bandwidths from 10~Gbps down to 80~Mbps, using the hardware and pipeline setup of Section~\ref{sec:experiment_setup}.
Figure~\ref{fig:throughput_gain} plots the resulting throughput gain of ResBM.
Even at 10~Gbps, ResBM is about $1.7\times$ faster (36.8k vs.\ 21.7k tokens per second), since the uncompressed activations still add communication time to every step.
As bandwidth drops, the uncompressed baseline slows sharply, whereas ResBM stays above 33k tokens per second down to 400~Mbps and still reaches 21.6k tokens per second at 80~Mbps.
The gain therefore grows to about $13\times$ at 1~Gbps, $16\times$ at 800~Mbps and $30\times$ at 400~Mbps, and reaches about $98\times$ at 80~Mbps, where the uncompressed baseline runs at only about 220 tokens per second.
ResBM's benefit thus grows as bandwidth falls, and it is substantial over the whole range we measured.

\subsection{Parameter, memory, compute and communication overhead}
\label{app:memory_overhead}

The bottleneck architecture introduces one encoder--decoder pair per pipeline boundary relative to the baseline.
With the shapes of Equation~\ref{eq:encoder_decoder_shapes} (encoder $H \to m \to h$, decoder $h \to m \to H$, $m = H/4$, no biases), a boundary adds $2\,(Hm + mh) = H^2/2 + Hh/2$ parameters.
For the 2B model ($H = 4096$, $h = 32$) spread across 8 pipeline stages with one transformer block per stage, each of the 7 boundaries adds 8,454,144 parameters, increasing the total parameter count from 1,896,751,104 to 1,955,930,112, an overhead of approximately 59.2M parameters (3.1\%).
At $100\times$ ($h = 40$), each boundary adds 8,470,528 parameters, for a total overhead of 59.3M parameters, also 3.1\%.

The added memory and compute are equally small.
Each intermediate stage holds one decoder and one encoder, 8.45M parameters at $128\times$.
With fp32 weights, gradients and AdamW states (16 bytes per parameter), these occupy about 135\,MB per stage, negligible next to the 24\,GB of the smallest GPUs in Appendix~\ref{app:isostage}.
The encoder and decoder are dense layers, so their training FLOPs per token scale with their parameters: counting 6 FLOPs per parameter per token for every weight except the input embedding, plus attention, they add about 3.4\% to the per-token training compute of the 2B baseline (0.36 of 10.35 GFLOPs). This is slightly more than their share of parameters, because the input embedding (0.2B parameters) is a lookup and costs no matrix multiplications.
Accordingly, ResBM's throughput matches SM's in Figures~\ref{fig:resbm_vs_all_high_comp}--\ref{fig:varying_split_count}.

The primary benefit lies in inter-stage communication.
Without bottleneck layers, each pipeline boundary transfers, for every sequence, an $L \times H = 1024 \times 4096$ activation tensor (8~MiB in bf16) per direction.
At $128\times$ compression, this reduces to $L \times h = 1024 \times 32$ (64~KiB in bf16) --- a $128\times$ reduction in per-boundary transfer volume for both forward activations and backward gradients.
Across the 7 stage boundaries in an 8-stage pipeline, and counting both directions, this translates from 112~MiB to 896~KiB per sequence; an optimizer step moves a further factor of the batch size on top of these figures.
Either way the ratio is unchanged, and it is what makes the architecture practical for low-bandwidth interconnects.

\clearpage
\section{Performance under practical infrastructure constraints}
\label{app:isostage}

Our main experiments (Section~\ref{sec:compress_local_perf}) deliberately place a
communication boundary after almost every transformer block, an extreme setting
chosen to stress compression. In practice, the number of pipeline stages and the
blocks assigned to each are set by the size and memory of the participating nodes
(Section~\ref{sec:practical_considerations}). This appendix evaluates ResBM in
such a configuration, taken from a real low-bandwidth training run.

Low-bandwidth pipeline parallelism targets compute that is not co-located:
smaller nodes spread across regions, consumer and workstation GPUs, or
preemptible spot capacity. In this setting, transformer blocks are assigned to
pipeline stages according to the memory of the available GPUs rather than to
balance FLOPs. The public 8B SM run of \citet{avraham2026agora}, trained over the
internet with Agora, the codebase that also provides our SM implementation, is a
representative example; we refer to it as Agora-8B. It splits 32 transformer blocks over
seven pipeline stages as $6\,|\,4{\times}5\,|\,6$: a first (head) stage of six
blocks, five intermediate (body) stages of four blocks each, and a last (tail)
stage of six blocks. As SM requires, the 26 blocks before the last communication
boundary are constrained to the shared subspace, and only the six blocks of the
tail operate at full rank. Each body fits a 24\,GB consumer GPU, whereas the
tail, which also holds the LM head, needs a GPU with 48\,GB or more. The pipeline
is therefore far from FLOP-balanced: the tail does nearly $3\times$ the FLOPs of
a body. The hardware absorbs this imbalance, because nodes are assigned to stages
by GPU memory and each stage can be served by many nodes.

We configure ResBM under the same constraints. It keeps the architecture of Agora-8B
(hidden width $H = 5120$, 40 attention heads, feed-forward width 18432, and a
128K-token vocabulary) and its seven pipeline stages, giving six communication
boundaries. It also keeps the per-token communication at each boundary: a
bottleneck width of $h = 52$ carries the same number of bytes as the 51 subspace
coefficients and the token id that Agora-8B transmits, corresponding to roughly
$100\times$ compression, the regime of our main experiments. Because ResBM adds an
encoder--decoder pair only at the six boundaries, all of its transformer blocks
are unconstrained. We choose the number of blocks per stage to match Agora-8B
stage by stage (Table~\ref{tab:isostage}):
\begin{itemize}
  \item The tail matches Agora-8B's, with six full-rank blocks and the LM head, so
        it targets the same GPU tier.
  \item The head matches Agora-8B's head in trainable parameters.
  \item The five bodies, with two or three blocks each, match Agora-8B's bodies in
        total trainable parameters, and each fits a 24\,GB GPU.
\end{itemize}
The resulting 21-block ResBM matches Agora-8B's trainable parameter count (8.66B
vs.\ 8.63B, $+0.3\%$). It needs less GPU memory on every stage, because it does
not keep a copy of SM's frozen token-embedding table on each stage. It also
uses 9\% fewer training FLOPs per token. We train ResBM and SM on FineWeb-Edu2 for 11B tokens,
with a batch size of 16 sequences and the network setup of
Section~\ref{sec:experiment_setup}. Both use the hyperparameters of the
Agora-8B run: AdamW with a peak learning rate of $3\times10^{-4}$, warmed up
linearly from $0.1\%$ of its peak over 4000 steps and then decayed linearly to
$1\%$ of its peak. We also train an uncompressed baseline with ResBM's 21
transformer blocks, split over seven pipeline stages, but without its
encoder--decoder pairs (8.58B trainable parameters).\footnote{All three models
run as synchronous pipelines in our trainer, without the asynchronous scheduling
of the original Agora-8B run. Owing to GPU availability, each runs as a single
pipeline replica on 7 H100 GPUs rather than on the heterogeneous GPUs of the
public run. The context length is 2048 tokens instead of 4096. ResBM and the
baseline use a cosine instead of a linear learning-rate decay. Both schedules decay over 1T
tokens, so the learning rate of every run stays within about $1\%$ of its peak.}

\begin{table}[ht]
\centering\small
\caption{Per-stage allocation of Agora-8B, which our SM run reproduces, and the
matched ResBM. c: block constrained to the shared subspace; f: full-rank block.
ResBM's head includes its trainable input embedding. Memory is estimated from
fp32 weights and gradients, with optimizer state offloaded to CPU (the default in
Agora), plus SM's frozen token-embedding table; activations are excluded. Totals
are computed before rounding.}
\label{tab:isostage}
\begin{tabular}{lcccccc}
\toprule
& \multicolumn{3}{c}{Agora-8B} & \multicolumn{3}{c}{ResBM} \\
\cmidrule(lr){2-4}\cmidrule(lr){5-7}
Stage & Blocks & Params & Mem. & Blocks & Params & Mem. \\
\midrule
Head        & 6c & 1.37B & 13.6\,GB & 2 & 1.36B & 10.9\,GB \\
Bodies 1--2 & 4c & 0.91B &  9.9\,GB & 2 & 0.71B &  5.7\,GB \\
Bodies 3--5 & 4c & 0.91B &  9.9\,GB & 3 & 1.05B &  8.4\,GB \\
Tail        & 6f & 2.73B & 24.5\,GB & 6 & 2.74B & 21.9\,GB \\
\midrule
Total       & 32 & 8.63B &          & 21 & 8.66B & \\
\bottomrule
\end{tabular}
\end{table}

As shown in Figure~\ref{fig:isostage}, ResBM overtakes SM within the first
billion tokens and keeps a lower loss for the rest of training, ending with a
loss gap to the uncompressed baseline of about $0.15$ nats, against $0.19$ for
SM. Over this budget, SM's gap does not clearly grow, unlike in the 2B
experiments of Section~\ref{sec:compress_local_perf}; longer runs are needed to
determine whether it does. ResBM also processes slightly more tokens per second
than SM (about 8.4k against 8.1k). This gain is smaller than its 9\% saving in
training FLOPs per token, because both models share the same tail, the most
expensive stage of the pipeline.

\clearpage
\section{Parameter-matched SM architectures (SM-M)}
\label{app:sm_m}

On the same backbone, SM trains fewer parameters than ResBM: it restricts the
out-projections of every block before the last pipeline boundary to the shared
subspace, and keeps its token embedding frozen. SM-M
(Section~\ref{sec:experiment_setup}) closes this gap by enlarging SM rather than
shrinking ResBM. We give its configuration for 2, 4 and 8 pipeline stages (1, 3
and 7 communication boundaries) at $100\times$ compression, where ResBM's
bottleneck width $h = 40$ carries the same bytes per token as SM's 39 subspace
coefficients plus the token id. For reference, the tables also give the
allocation of SM itself.

SM-M keeps SM's constraint: every block before the last boundary is
subspace-constrained, and only the blocks of the last stage are full-rank. Since a constrained block costs fewer FLOPs than a full-rank one, we add
constrained blocks and distribute them to balance training FLOPs across stages,
as closely as integer block counts allow, so SM-M holds more blocks per stage
than ResBM. We then set SM-M's feed-forward width, within about 4\% of ResBM's
11008, so that its trainable parameter count matches ResBM's to within 0.25\%.
Tables~\ref{tab:sm_m_arch_pp2}--\ref{tab:sm_m_arch} give the per-stage
allocations. Because of its additional blocks, SM-M uses 13.3--14.3\% more
training FLOPs per token than ResBM.

\begin{table}[ht]
\centering\small
\setlength{\tabcolsep}{3.5pt}
\caption{Per-stage allocation of SM, SM-M and ResBM with 2 pipeline stages
(1 boundary). SM-M uses feed-forward width 10624 (ResBM: 11008), giving
$-0.23\%$ trainable parameters and $+13.3\%$ training FLOPs per token relative
to ResBM. c: block constrained to the shared subspace; f: full-rank block; all
ResBM blocks are unconstrained; ResBM's head includes its trainable token
embedding, and every tail includes the LM head. Memory is estimated from fp32 weights and gradients, with optimizer
state offloaded to CPU (the default in Agora), plus the frozen token-embedding
table of SM and SM-M (0.8\,GB), held on every stage; activations are excluded.}
\label{tab:sm_m_arch_pp2}
\begin{tabular}{lccccccccc}
\toprule
& \multicolumn{3}{c}{SM} & \multicolumn{3}{c}{SM-M} & \multicolumn{3}{c}{ResBM} \\
\cmidrule(lr){2-4}\cmidrule(lr){5-7}\cmidrule(lr){8-10}
Stage & Blocks & Params & Mem. & Blocks & Params & Mem. & Blocks & Params & Mem. \\
\midrule
Head  & 4c & 499M   & 4.8\,GB & 8c & 971M   & 8.6\,GB & 4 & 953M   & 7.6\,GB \\
Tail  & 4f & 948M   & 8.4\,GB & 4f & 929M   & 8.3\,GB & 4 & 953M   & 7.6\,GB \\
\midrule
Total & 8  & 1.448B &               & 12 & 1.901B &         & 8 & 1.905B & \\
\bottomrule
\end{tabular}
\end{table}

\begin{table}[ht]
\centering\small
\setlength{\tabcolsep}{3.5pt}
\caption{Per-stage allocation of SM, SM-M and ResBM with 4 pipeline stages
(3 boundaries). SM-M uses feed-forward width 11328 (ResBM: 11008), giving
$-0.06\%$ trainable parameters and $+13.9\%$ training FLOPs per token relative
to ResBM. Body rows are per stage. Notation and memory as in
Table~\ref{tab:sm_m_arch_pp2}.}
\label{tab:sm_m_arch_pp4}
\begin{tabular}{lccccccccc}
\toprule
& \multicolumn{3}{c}{SM} & \multicolumn{3}{c}{SM-M} & \multicolumn{3}{c}{ResBM} \\
\cmidrule(lr){2-4}\cmidrule(lr){5-7}\cmidrule(lr){8-10}
Stage & Blocks & Params & Mem. & Blocks & Params & Mem. & Blocks & Params & Mem. \\
\midrule
Head        & 1c & 126M   & 1.8\,GB & 4c & 510M   & 4.9\,GB & 1 & 396M   & 3.2\,GB \\
Bodies 1--2 & 3c & 373M   & 3.8\,GB & 4c & 508M   & 4.9\,GB & 3 & 565M   & 4.5\,GB \\
Tail        & 1f & 392M   & 4.0\,GB & 1f & 395M   & 4.0\,GB & 1 & 396M   & 3.2\,GB \\
\midrule
Total       & 8  & 1.264B &               & 13 & 1.921B &         & 8 & 1.922B & \\
\bottomrule
\end{tabular}
\end{table}

\begin{table}[ht]
\centering\small
\setlength{\tabcolsep}{3.5pt}
\caption{Per-stage allocation of SM, SM-M and ResBM with 8 pipeline stages
(7 boundaries), the default setting of Section~\ref{sec:experiment_setup}. SM-M
uses feed-forward width 10560 (ResBM: 11008), giving $+0.01\%$ trainable
parameters and $+14.3\%$ training FLOPs per token relative to ResBM. Body rows
are per stage. Notation and memory as in Table~\ref{tab:sm_m_arch_pp2}.}
\label{tab:sm_m_arch}
\begin{tabular}{lccccccccc}
\toprule
& \multicolumn{3}{c}{SM} & \multicolumn{3}{c}{SM-M} & \multicolumn{3}{c}{ResBM} \\
\cmidrule(lr){2-4}\cmidrule(lr){5-7}\cmidrule(lr){8-10}
Stage & Blocks & Params & Mem. & Blocks & Params & Mem. & Blocks & Params & Mem. \\
\midrule
Head        & 1c & 126M   & 1.8\,GB & 1c & 123M   & 1.8\,GB & 1 & 396M   & 3.2\,GB \\
Bodies 1--6 & 1c & 124M   & 1.8\,GB & 2c & 241M   & 2.8\,GB & 1 & 194M   & 1.6\,GB \\
Tail        & 1f & 392M   & 4.0\,GB & 1f & 386M   & 3.9\,GB & 1 & 396M   & 3.2\,GB \\
\midrule
Total       & 8  & 1.264B &               & 14 & 1.956B &         & 8 & 1.956B & \\
\bottomrule
\end{tabular}
\end{table}

\end{document}